\documentclass[11pt,a4paper]{article}
\usepackage[margin=2.9cm,top=2.5cm,bottom=2.5cm]{geometry}

\usepackage{authblk}

\usepackage[utf8]{inputenc} 
\usepackage[T1]{fontenc}    
\usepackage[english]{babel}
\usepackage{xargs}

\usepackage{lmodern}
\usepackage{montserrat}
\usepackage{fourier}
\DeclareMathAlphabet{\mathcal}{OMS}{cmsy}{m}{n}

\usepackage[sf,raggedright]{titlesec}
\titleformat{\paragraph}[runin]{\rmfamily\bfseries}{}{0em}{}[.]
\titleformat{\section}{}{\sffamily\Large\bfseries\thetitle}{0.8em}{\sffamily\Large\bfseries}
\titleformat{\subsection}{}{\sffamily\large\bfseries\thetitle}{0.5em}{\sffamily\large\bfseries}
\titleformat{\subsubsection}{}{\sffamily\large\bfseries\thetitle}{0.5em}{\sffamily\large\bfseries}
\titlespacing{\paragraph}{0pt}{0.5em}{6pt}
\titlespacing{\subsection}{0pt}{1em}{0.5em}
\titlespacing{\subsubsection}{0pt}{0.8em}{0.5em}

\usepackage{graphicx}
\usepackage[dvipsnames]{xcolor}
\usepackage[colorlinks = true, citecolor = blue,linkbordercolor=black,linkcolor=red]{hyperref}
\usepackage[inline]{enumitem}
\usepackage{aliascnt}
\usepackage{wrapfig}

\usepackage[square,numbers,sort]{natbib}
\usepackage[tbtags]{amsmath}
\usepackage{amsthm}
\usepackage{bm}
\usepackage{amssymb,mathrsfs}
\usepackage{amsfonts}
\usepackage{mathtools}
\usepackage{nicefrac}       

\usepackage{todonotes}

\usepackage{cleveref} 

\allowdisplaybreaks
\numberwithin{equation}{section}

\definecolor{JungleGreen}{RGB}{40, 169, 143}
\definecolor{brightmaroon}{rgb}{0.76, 0.13, 0.28}
\definecolor{BatteryChargedBlue}{RGB}{42, 187, 218}
\definecolor{GargoyleGas}{RGB}{250, 213, 66}
\definecolor{LightCarminePink}{RGB}{236, 94, 100}
\definecolor{MaximumPurple}{RGB}{107, 57, 121}

\definecolor{lightred}{rgb}{1, 0.8, 0.8}

\makeatletter
\newcommand{\Josua}[2][]{\@todo[color=BatteryChargedBlue, #1]{Josua: #2}}
\newcommand{\Samuel}[2][]{\@todo[color=BatteryChargedBlue, #1]{Samuel: #2}}
\makeatother

\makeatletter
\newtheorem{theorem}{Theorem}
\crefname{theorem}{theorem}{Theorems}
\Crefname{Theorem}{Theorem}{Theorems}

\newaliascnt{corollary}{theorem}

\aliascntresetthe{corollary}
\crefname{corollary}{corollary}{corollaries}
\Crefname{Corollary}{Corollary}{Corollaries}

\newaliascnt{proposition}{theorem}
\newtheorem{proposition}[proposition]{Proposition}
\aliascntresetthe{proposition}
\crefname{proposition}{proposition}{propositions}
\Crefname{Proposition}{Proposition}{Propositions}

\newaliascnt{definition}{theorem}
\newtheorem{definition}[definition]{Definition}
\aliascntresetthe{definition}
\crefname{definition}{definition}{definitions}
\Crefname{Definition}{Definition}{Definitions}

\newaliascnt{remark}{theorem}
\newtheorem{remark}[remark]{Remark}
\aliascntresetthe{remark}
\crefname{remark}{remark}{remarks}
\Crefname{Remark}{Remark}{Remarks}

\newtheorem{example}[theorem]{Example}
\crefname{example}{example}{examples}
\Crefname{Example}{Example}{Examples}

\crefname{figure}{figure}{figures}
\Crefname{Figure}{Figure}{Figures}

\newcommand{\tr}{\mathrm{tr}}

\newcommand{\manifold}{\mathcal{M}}

\newcommand{\metric}{{\mathfrak{g}}}
\newcommand{\covDeriv}{\frac{D}{dt}}
\newcommand{\vecFields}{C^\infty(\mathcal{M},T\mathcal{M})}
\newcommand{\parTp}{\mathfrak{T}}

\newcommand{\point}{p}

\newcommand{\chart}{\psi}
\newcommand{\coord}{\xi}
\newcommand{\curve}{\gamma}
\newcommand{\vf}{V}
\newcommand{\wf}{W}

\newcommand{\pathenergy}{\mathcal{E}}
\newcommand{\length}{\mathcal{L}}

\newcommand{\Hamiltonian}{H}
\newcommand{\momentum}{p}

\newcommand{\Sym}{\mathcal{S}}
\newcommand{\SPD}{\Sym_{++}}
\newcommand{\Grassmann}{\mathrm{Gr}}
\newcommand{\Stiefel}{\mathrm{St}}
\newcommand{\Imm}{\mathrm{Imm}}
\newcommand{\Wasserstein}{W}

\newcommand{\argmin}{\operatorname{argmin}}

\newcommand{\loss}{\mathfrak{L}}

\def\mcx{\mathcal{X}}
\def\dist{\mathbf{d}}
\def\ie{\textit{i.e.}}
\newcommand{\R}{\mathbb{R}}
\newcommand{\Z}{\mathbb{Z}}
\def\eg{e.g.}
\newcommand*{\dd}{\mathop{}\!\mathrm{d}}
\def\Id{\mathrm{Id}}
\def\Idd{\mathrm{I}_d}
\def\Exp{\mathrm{Exp}}
\def\Log{\mathrm{Log}}
\def\vol{\text{vol}}
\def\msx{\mathsf{X}}
\def\diag{\operatorname{diag}}

\def\iid{i.i.d.}

\def\eqsp{\;}

\title{A Review on Riemannian Metric Learning: Closer to You than You Imagine}

\author[1]{Samuel Gruffaz \thanks{samuel.gruffaz@ens-paris-saclay.fr}} 
\author[1]{Josua Sassen \thanks{josua.sassen@ens-paris-saclay.fr}}

\affil[1]{Université Paris-Saclay, ENS Paris-Saclay, Centre Borelli, F-91190 Gif-sur-Yvette, France.}

\date{}

\begin{document}

\maketitle

\begin{abstract}
    Riemannian metric learning is an emerging field in machine learning, unlocking new ways to encode complex data structures beyond traditional distance metric learning.
    While classical approaches rely on global distances in Euclidean space, they often fall short in capturing intrinsic data geometry. 
    Enter Riemannian metric learning: a powerful generalization that leverages differential geometry to model the data according to their underlying Riemannian manifold. 
    This approach has demonstrated remarkable success across diverse domains, from causal inference and optimal transport to generative modeling and representation learning. 
    In this review, we bridge the gap between classical metric learning and Riemannian geometry, providing a structured and accessible overview of key methods, applications, and recent advances. 
    We argue that Riemannian metric learning is not merely a technical refinement but a fundamental shift in how we think about data representations. 
    Thus, this review should serve as a valuable resource for researchers and practitioners interested in exploring Riemannian metric learning and convince them that it is closer to them than they might imagine---both in theory and in practice.
\end{abstract}

\section{Introduction}

In recent decades, machine learning research has focused on developing vector-based representations for various types of data, including images, text, and time series \cite{bengio2013representation}.  
Learning a meaningful representation space is a foundational task that accelerates research progress, as exemplified by the success of Large Language Models (LLMs) \cite{vaswani2017attention}.  
A complementary challenge is learning a distance function (defining a metric space) that encodes aspects of the data's internal structure.
This task is known as \textit{distance metric learning}, or simply \textit{metric learning} \cite{bellet2015metric}.  
Metric learning methods find applications in every field using algorithms relying on a distance such as the ubiquitous k-nearest neighbors classifier:
Classification and clustering \cite{xing2002distance}, recommendation systems \cite{hsieh2017collaborative}, optimal transport \cite{cuturi2014ground}, and dimension reduction \cite{lin2008riemannian,wang2015survey}.  
However, when using only a global distance, the set of available modeling tools to derive computational algorithms is limited and does not capture the intrinsic data structure.

Hence, in this paper, we present a literature review of \textit{Riemannian metric learning}, a generalization of metric learning that has recently demonstrated success across diverse applications, from causal inference \cite{dominguez2023data,pradier2024beyond,farzam2024geometry} to generative modeling \cite{lavenant2024toward,kapusniak2024metric,sun2025geometry}.  
Unlike metric learning, \textit{Riemannian} metric learning does not merely learn an embedding capturing distance information, but estimates a Riemannian metric characterizing distributions, curvature, and distances in the dataset, \ie~the Riemannian structure of the data.
This naturally leads to modeling tools inspired by Riemannian geometry, such as geodesics \cite{lavenant2024toward,kapusniak2024metric,sun2025geometry} and Ricci curvature \cite{farzam2024geometry}, further expanding the scope of geometric approaches for addressing modeling and representation learning problems. 
Thus, we argue that Riemannian metric learning is a fundamental task in representation learning, enriching applications by leveraging geometric theory---much like generative modeling has benefited heavily from probability theory \cite{ho2020denoising}.

\begin{figure}[t]
	\includegraphics[width=130mm]{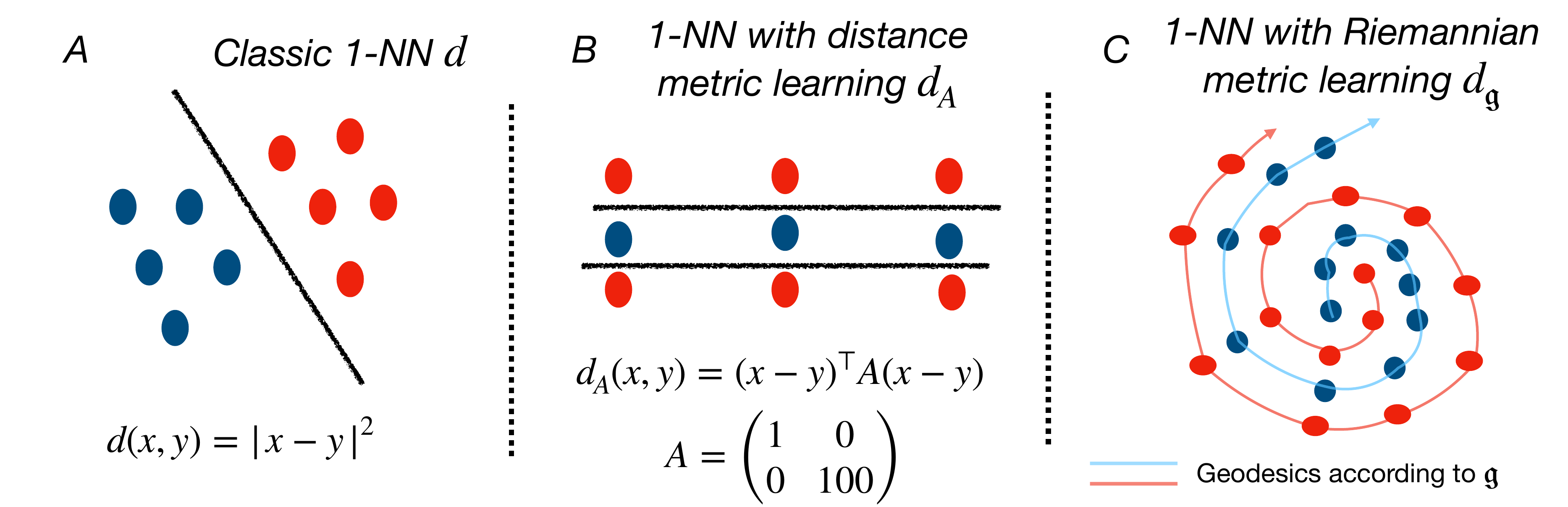}
	\centering
	\caption{
		On the left, a classic 1-NN using euclidean distance can separate red and blue points. 
		In the middle, a 1-NN using distance metric learning can separate the points on the anisotropic grid, but not a classic 1-NN. 
		On the right, only a 1-NN using Riemannian metric learning can achieve the separation of spiral data.}
	\label{fig:classifciation}
\end{figure}

To recount the story from the beginning, let us begin with the principles of metric learning.
Metric learning \cite{bellet2015metric} is an integral, though often implicit, aspect of machine learning.
Whether consciously or not, most practitioners in the field are already engaging with it---or will eventually.
The concept has a long history, dating back to early methods like the k-nearest neighbors (k-NN) classifier \cite{cover1967nearest,short1981optimal,friedman1994flexible}.
As the term `nearest' suggests, a measure of distance is at the heart of k-NN algorithms as depicted in \Cref{fig:classifciation}.
Without prior knowledge, the distance is often selected as the Euclidean distance.
However, this choice is uninformative of the input and output distribution.
That is why Atkeson et al.~\cite{atkeson1997locally} suggested to learn the distance by learning a symmetric positive definite matrix $G\in \R^{d\times d}$ such that $\dist_G(x,y)=\sqrt{(y-x)^\top G(y-x)}$.
It turns out that learning $G=L^\top L$ is equivalent to learning a linear transformation $x \mapsto Lx$ of the input space as $\dist_G=\dist(L\cdot,L\cdot)$.
The matrix \( G \) can be learned, for example, in an unsupervised manner using Principal Component Analysis (PCA) or in a supervised manner using Linear Discriminant Analysis (LDA).
Learning \( G \) is valuable for two key reasons:
First, it assigns \textit{global relative weights to the different coordinates} when computing distances.
Second, it allows us to account for \textit{global correlations} in the dataset. 
However, \( G \) needs to vary to account for \textit{heterogeneous local geometries} within the dataset as shown in \Cref{fig:classifciation}.
This linear framework was then extended with non-linear transformations $x \mapsto \phi(x)$ by using kernels or neural networks \cite{yang2006distance,duan2018deep,ghojogh2022spectral} such that the distance is $\dist(\phi(\cdot),\phi(\cdot))$, which reduces to learning a feature function $\phi$ as already pointed out.

Going one step further, metric learning can be embedded in a more general framework called Riemannian metric learning \cite{peltonen2004improved}, in the same way that Galilean relativity is embedded by General relativity \cite{einstein1905elektrodynamik}. 
In Riemannian metric learning, the input space $(\mcx,\metric)$ is a Riemannian manifold, \ie~a kind of smooth geometric shape (see \Cref{fig:preli_scheme_tangent_space}) with a Riemannian metric $\metric \colon \mcx\to \R^{d\times d} $ that allows to locally measure lengths and angles via a matrix field.
From this local geometric structure, one can define a global distance between any $x,y\in \mcx$ as
\begin{equation}
	\label{eq:dist_as_geodesics}
	\dist(x,y)=\inf_{\gamma\in C^1([0,1],\mcx), \gamma(0)=x, \gamma(1)=y} \int_0^1 \sqrt{\dot{\gamma}(t)^\top \metric(\gamma(t))\dot{\gamma}(t)} \eqsp .
\end{equation}
Minimizing curves of \eqref{eq:dist_as_geodesics} are called metric geodesics and can be understood as generalizations of straight lines to manifolds as illustrated in \Cref{fig:classifciation}.
If the Riemannian metric is constant and the manifold is the Euclidean space, then geodesics are straight lines and we recover the distance metric learning framework from above.
However, Riemannian metric learning is not just a generalization of distance metric learning with non-linear transformations; it also gives new mathematical objects useful for modeling.
For instance, geodesics can express trajectories for longitudinal modeling and interpolation \cite{louis2019riemannian,kushnerlearning,kapusniak2024metric}, parallel transport (defined later) can be used in transfer learning \cite{tam2023transfer,guigui2022parallel}, and the metric volume can be used for modeling probability distribution and sampling \cite{cui2024optimal,dekruiff2024pullback}.
All these mathematical objects are related to each other and have been extensively studied in Riemannian Geometry \cite{do1992riemannian,sommer2020introduction}. Riemannian metric \textit{learning} leverages the wealth of this literature to offer both flexible and interpretable modeling tools.

As all these mathematical objects depend on the Riemannian metric $\metric$, it is worthwhile to learn it by minimizing an objective function.
First, as mentioned previously, the metric was chosen constant in an Euclidean space \cite{xing2002distance,davis2007information,zadeh2016geometric} and the objective function asked to reduce the distance between similar points and increase the distance between dissimilar points.
The optimization was often convex and was solved using gradient descents, semi-definite programming or Support Vector Machine solvers \cite{wang2014kernel,bellet2013survey}.
Then, people tried to learn metrics on other types of manifolds such as the Grassmanian, the manifold of SPD matrices \cite{zhu2018towards}, and the manifold of probability distributions \cite{cuturi2014ground}.
Furthermore, new parameterizations of the metric have been suggested to handle more complex geometry based on locally constant approximations \cite{qiu2024estimating}, kernels \cite{arvanitidis2016locally,gruffaz2021learning}, and neural networks \cite{duan2018deep,chadebec2020geometry,kalatzis2020variational}.
The optimization is highly non-convex and relies sometimes on sophisticated techniques \cite{benmansour2010derivatives,cui2024optimal,sun2025geometry}, although often standard accelerated gradient descent techniques proof to be sufficient.

\paragraph{Contributions}
There exists a wide range of reviews on \textit{distance} metric learning \cite{yang2006distance,bellet2013survey,kulis2013metric,wang2015survey,duan2018deep,suarez2021tutorial,ghojogh2022spectral}:
Yang and Jin~\cite{yang2006distance} provided the first, then Bellet et al.~\cite{bellet2013survey} and Kulis et al.~\cite{kulis2013metric} offered an interesting classification of metric learning methods, Wang and Sun~\cite{wang2015survey} focused on dimension reduction methods while Duan et al.~\cite{duan2018deep} addressed deep learning methods, Suárez et al.~\cite{suarez2021tutorial} distinguished themselves from the others by their comprehensive benchmark, and finally Ghojogh et al.~\cite{ghojogh2022spectral} exposed the different mathematical formalism exhaustively.
However, none of them tackles \textit{Riemannian} metric learning.
Recently, Riemannian metric learning has impacted popular research topics from generative modeling \cite{lavenant2024toward,kapusniak2024metric,sun2025geometry} to causal inference \cite{dominguez2023data,pradier2024beyond,farzam2024geometry}.
It was crucial to improve the state-of-the-art considerably both in performance and interpretability. 
That is why, considering the diversity of publications in Riemannian metric learning and their rising impact, we dedicate our review to this topic:
\begin{itemize}
    \item We present Riemannian metric learning as a generalization of distance metric learning and state its general framework.
    \item We motivate this framework by highlighting the main applications arising from different research communities, demonstrating the wealth and interest of the Riemannian formalism.
    \item We offer a classification of Riemannian metric learning methods according to their metric parametrizations, related computational methods, objectives, and optimization methods.
 The key points of distance metric learning will also be recalled in this context to give a complete picture. 
    \item We propose promising theoretical and practical research directions for Riemannian metric learning.
\end{itemize}

\paragraph{Structure}
First, we recall preliminary concepts on the essential tools and properties of Riemannian geometry in \Cref{sec:preliminaries}; readers familiar with these notions may safely skip this section. 
At the same time, we introduce the main manifolds relevant to the data used in Riemannian metric learning in \Cref{sec:typical_manifold}, along with the manifold of Riemannian metrics, to provide a broader perspective on the subject.
Second, the formal optimization problem associated with Riemannian metric learning is presented in \Cref{sec:abstract_problem} to define the topic rigorously. 
This is followed by an overview of its numerous applications in \Cref{sec:applications}, highlighting the practical benefits and key techniques of Riemannian formalism.
Third, we discuss how to perform Riemannian metric learning in \Cref{sec:how}, classifying the methods based on their metric parameterization (\Cref{sec:parametric}) and associated computational approaches (\Cref{sec:computing}), their objectives (\Cref{sec:objectives}), and their optimization strategies (\Cref{sec:optimization}). 
This classification helps clarify the various technical challenges that arise when designing Riemannian metric learning methods.
Finally, before concluding in \Cref{sec:conclusion}, we outline key research directions in Riemannian metric learning in \Cref{sec:future}, which may inspire both theorists and practitioners in the applied mathematics community.

\paragraph{Acknowledgments}
This project has received funding from the European Union's Horizon 2020 research and innovation program under the Marie Skłodowska-Curie grant agreement No\\ 101034255.

\subsection{Notation}
We denote by $\mathcal{P}(\msx)$ the power set of a set $\msx$, integer ranges by $[k:l]=\{k,\ldots,l\}\subset \mathcal{P}(\Z)$ and $ [l]=[1:l]$ with $k,l\in \mathbb{N}$.
  The canonical scalar product of two vectors $x=(x_i)_{i\in[d]},y=(y_i)_{i\in[d]}\in \R^d$ is denoted by $x^\top y =\sum_{i=1}^d x_i y_i$. 
  The Euclidean norm is denoted by $||\cdot||=\sqrt{\cdot^\top \cdot}$.
  We denote by $\mathrm{C}^k(\mathsf{U}, \R^{d})$ the set of all $k$ times continuously differentiable functions from $\mathsf{U}$ to $\R^{d}$.
  When $\mathsf{U}\subset \R$, we denote by $\dot\gamma$ the time derivative of $\gamma\in \mathrm{C}^1(\mathsf{U}, \R^{d})$.
 The spaces of square symmetric and symmetric positive definite matrices are denoted by $\mathcal{S}^d=\{A\in \R^{d\times d} \, : \, A^{\top} = A\}$ and $\mathcal{S}^d_{++}=\{A\in \mathcal{S}^d \, : x^\top Ax > 0\, \text{for any} \, x\in\R^d \setminus \{0\}\}$, respectively.
 The square identity matrix in dimension $d\times d$ is denoted by $\Idd$. 
 The determinant and the trace of a matrix $A\in \R^{d\times d}$ are denoted by $\tr(A)$ and $\det(A)$ respectively.
  For any $(x_i)_{i\in[m]}\in \R^m$,  $\diag((x_i)_{i\in[m]})$ denotes the diagonal matrix whose diagonal elements are $(x_i)_{i\in[m]}$, arranged from the top-left.
  For any $f\in  \mathrm{C}^1(\mathsf{U}, \R^{m})$ with $\mathsf{U}\subset \R^{d}$, $\dd f$ is the differential of $f$. 
  For any $f\in  \mathrm{C}^1(\mathsf{U}, \R)$ with $\mathsf{U}\subset \R^{d}$, $\nabla f$ is the gradient of $f$, but can also denotes the covariate derivative depending on the context.
  For any $j\in [d]$ and any $\chart \in  \mathrm{C}^1(\mathsf{U}, \R^{m})$ with $\mathsf{U}\subset \R^{d}$,  $\partial_{j} \chart \in \R^m$ denotes the directional directive of $\chart$ in direction $e_j$, where $e_j$ is the $j$-th element of the canonical basis.
  When a random variable $X\in E$ follows a probability distribution $P$ on a measurable space $(E,\mathcal{B})$, we write $X\sim P$.
  $\mathcal{U}([0,1])$ denotes the uniform probability distribution on $[0,1]$.
   If $P_1$ and $P_2$ are two probability distribution on $(E,\mathcal{B}_1)$ and $(F,\mathcal{B}_2)$ respectively, then $P_1 \otimes P_2$ denotes their product measure.

\section{A Primer on Riemannian Geometry} 
\label{sec:preliminaries}

In metric learning, the core element is naturally the metric — a measure of distance within the data space $\mathcal{X}$, which can be tailored to various objectives.
We begin by introducing the formal definition of a metric space:
\begin{definition}[Metric space]
    \label{def:metric_space}
    A \emph{metric space} consists of a set \(\mathcal{X}\) and a function \(\dist\colon \mathcal{X} \times \mathcal{X} \to \R\) called the \emph{metric}, such that
    \begin{enumerate}
        \item \(\dist(x, y) = 0\) if and only if \(x=y\),
        \item \(\dist(x, y) = \dist(y,x) \), and
        \item \(\dist(x, z) \leq \dist(x,y) + \dist(y,x)\).
    \end{enumerate}
\end{definition}
On Euclidean spaces, \ie~\(\mathcal{X} = \R^d\), the standard metric is the \emph{Euclidean distance}, given by \(\dist(x,y) = \lVert x - y \rVert_{\R^d} = \sqrt{\sum_{i=1}^d (x_i - y_i)^2}\).
However, depending on the application, alternative metrics such as the \emph{Manhattan distance}, defined as \(\dist(x,y) = \sum_{i=1}^d \lvert x_i - y_i\rvert\), are also employed.

Given some objective function, the metric also can be defined as the solution of an optimization problem, parametrizing the metric by $G\in \SPD^d$ as \(\dist_G(x,y) = \sqrt{(x-y)^T G (x-y)}\) for vectors \(x,y \in \R^d\).
This specific parametric form has been the dominant approach in most works on \emph{distance metric learning}.
However, such metrics are constant across the entire space and may fail to capture the inherent nonlinear structure of the data, similar to how linear PCA may not provide a meaningful approximation in the presence of complex data structures.
This limitation has motivated researchers to explore manifold learning methods.

In this work, we aim to approach metric learning from a Riemannian perspective. 
Therefore, we will briefly introduce key concepts from \emph{Riemannian geometry} necessary to understand the remainder of this review.
This section is essentially pedagogical and readers already familiar in Riemannian geometry can safely skip this part.
For a more comprehensive treatment, we refer the reader to standard textbooks such as \cite{lee2006riemannian}.

\begin{figure}[t]
    \includegraphics[width=130mm]{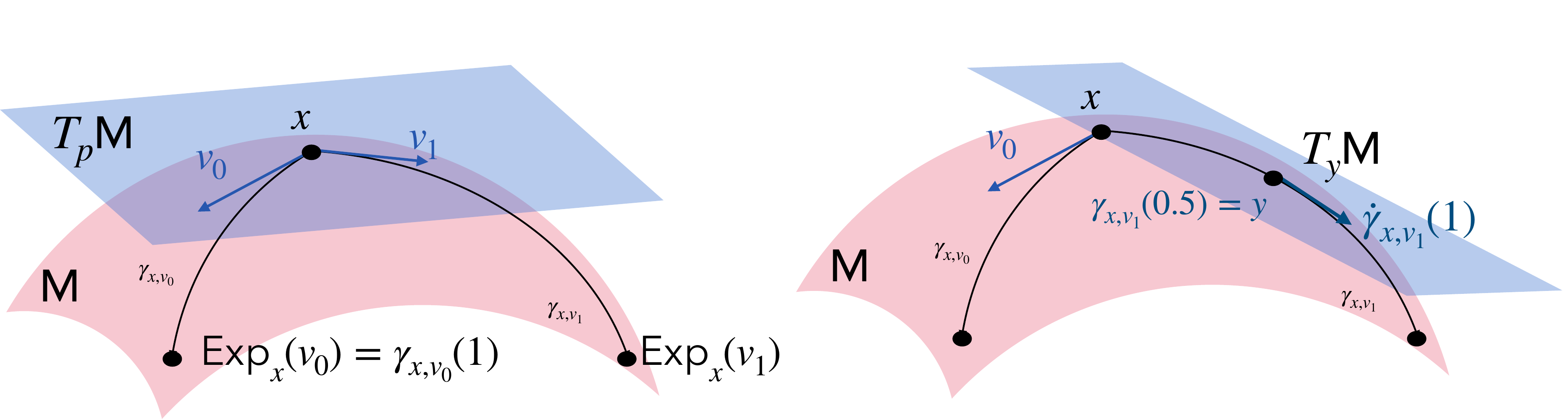}
    \centering
    \caption{
    	Illustration of tangent spaces, geodesics and the Exponential map.
    	The curve $\gamma_{x,v}$ is the geodesic starting from $x\in \manifold$ with velocity $v\in T_x\manifold $.
    }
    \label{fig:preli_scheme_tangent_space}
\end{figure}

\paragraph{Manifolds}
A \emph{differentiable manifold} \(\manifold\subset \R^{d_a}\) is formally defined in \Cref{sec:appendix_extended_preli} or in \cite[Chapter 2]{lee2006riemannian}, where \(d_a\) denotes the dimension of the ambient space.
For a first reading, and to simplify notation\footnote{All concepts introduced below are compatible with changes of parametrization in overlapping regions and can therefore be generalized to manifolds requiring multiple parametrizations.}, a \emph{differentiable manifold} of dimension \(d < \infty\) can be understood as the image of a global, surjective, and smooth parametrization \(\chart \colon U \subset \R^d \to \manifold\), where \(U\) is an open subset of \(\R^d\).
In general, we would need more than one parametrization and all concepts presented hereafter generalize to this scenario.
Readers unfamiliar with this concept may safely think of a differentiable manifold as a smoothly curved, nonlinear object, such as a curved surface in three-dimensional space as depicted in \Cref{fig:preli_scheme_tangent_space}.
Note that in all the following a function $f\colon\manifold\to \mathcal{Y}\subset \R^m$ is called smooth or infinitely differentiable if $ f\circ \psi \colon U\to \mathcal{Y}$ is so, which offers a valid definition since $U$ is an open set of $\R^d$. 

With this understanding, we can examine a local linearization of this curved space to apply tools from Euclidean vector spaces, which leads to the concept of a tangent space:
\begin{definition}[Tangent space]
    \label{def:general_tangent_space}
    The \emph{tangent space} \(T_\point \manifold\) of \(\manifold\) at \(\point \in \manifold\) is defined as 
    \begin{equation}
    	T_\point \manifold \coloneqq \{ \dot \curve(0) \mid \curve \in \mathrm{C}^1(-\varepsilon, \varepsilon, \manifold),\, \curve(0) = p, \varepsilon > 0 \},
    \end{equation}
    which is a vector space of dimension \(d\). 
    If \(\chart \colon U \subset \R^d \to \manifold \) is a parametrization with \(\chart(\coord) = \point\) for some \(\coord \in U\), then \(\partial_{\coord_i} \chart\) for \(i=1,\cdots,d\) is a basis of  \(T_\point \manifold\), called the \emph{canonical basis}.
    
\end{definition}
The dimension of the manifold is thus given by the dimension of possible local displacements in each point.
In the case of two-dimensional surfaces, the tangent space is simply the plane that is tangent to the surface at a given point as illustrated in \Cref{fig:preli_scheme_tangent_space}.
 
\begin{example}
	The Euclidean space $\R^d$ is a manifold of dimension $d$. 
	For any $x\in \R^d$, the tangent space is isomorphic to the the Euclidean space $T_x\R^d \simeq \R^d$ itself.
	Despite its simplicity, $\manifold=\R^d$ is still interesting in many applications when endowed with a spatially varying Riemannian metric.
	
	A non-trivial manifold prevalent in statistics is the manifold of symmetric positive definite (SPD) matrices 
	\begin{equation}
	    \label{eq:SPD}
	    \SPD^d \coloneqq \left\{M \in \R^{d\times d} \mid M = M^T \text{ and } x^TMx > 0 \text{ for any }x \in \R^d \backslash \{0\} \right\}.
	\end{equation}
	For any $M\in \SPD^d$, the tangent space $T_M\SPD^d$ is isomorphic to the space of symmetric matrices $\mathcal{S}^d$.
	Thus, $\SPD^d$ is a manifold of dimension $d(d+1)/2$.
\end{example}
More manifold examples and insights related to computations and modeling will be given in \Cref{sec:typical_manifold}.

\paragraph{Riemannian Structure}
Next, we aim to measure distances on our curved space, which requires introducing additional geometric structure.
Specifically, we will define this structure locally on tangent spaces and then use it to derive global notions of distance and related concepts across the entire manifold.
As previously discussed, a scalar product allows us to measure lengths and angles. 
Therefore, we begin by defining a scalar product for each tangent space:
\begin{definition}[Riemannian manifold]
    Let \(\manifold\) be a \(d\)-dimensional differentiable manifold.
    A \emph{Riemannian metric} on \(\manifold\) is a family of bilinear, symmetric and positive-definite  forms 
    \(\metric_\point \colon T_\point \manifold \times T_\point \manifold \to \R \) smoothly varying with \(\point \in \manifold\), in the sense that for a parametrization  \(\chart \colon U \subset \R^d \to \manifold \) the map \(\coord \mapsto \metric_{ij}(\xi)\coloneqq \metric_{\chart(\xi)}(\partial_{i} \chart(\xi), \partial_{j} \chart(\xi))\) is a smooth function on \(U\).
    A manifold equipped with a Riemannian metric is called a \emph{Riemannian manifold}.
\end{definition}
We can also interpret this inner product as a smooth matrix field \(\metric\colon\manifold\to \R^{d\times d}\) on the manifold by selecting a basis for each tangent space.
In the case that a global parametrization of the manifold exists, this corresponds to a smooth matrix field \(\metric\colon\R^d\to \R^{d\times d}\) defined on the parameter domain, representing a locally varying notion of distance.
Since \((\metric_{ij})_{ij}\) is an invertible matrix in \(\R^{d\times d}\), it has an inverse \(\metric^{-1} \in \R^{d\times d}\), which we denote as \((\metric^{kl})_{kl}\), satisfying \(\metric_{ij}\metric^{jk} = \delta_{ik}\).
\begin{example}
	\label{ex:affine_invariant_metric}
	The Euclidean scalar product on $\SPD^d$, \ie\,  $\metric_M(W,V)=\tr(W^\top V)$ for any $M\in\SPD^d$ and $W,V\in \mathcal{S}^d$, yields a Riemannian metric.
	However, this does not penalize displacements toward symmetric matrices with nonpositive eigenvalues.
	Hence, it does not take the structure of the manifold into account.
	An alteranative choice of Riemannian metric penalizing singularities is the affine-invariant metric $\metric_M(W,V)=\tr(M^{-1}WM^{-1}V)$ which implies that $\metric_M(V,V)$ diverges towards infinity when $M$ converges to the boundary of $\SPD^d$.
\end{example}

\begin{figure}[t]
    \includegraphics[width=130mm]{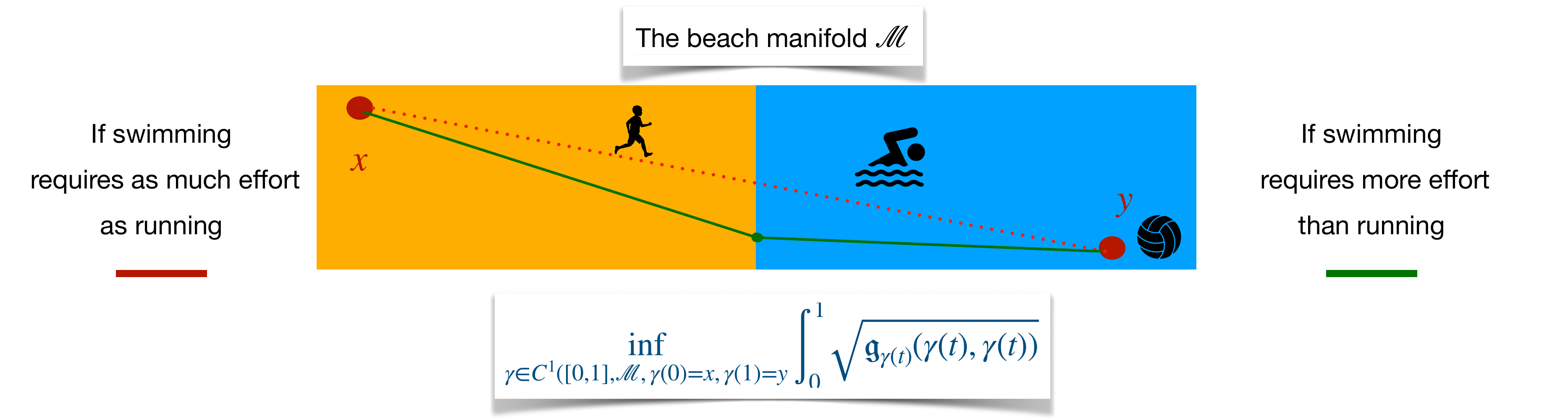}
    \centering
    \caption{Effect of the metric on the shape of the minimizing curve in \eqref{eq:dist_opt}.
     The effort required for displacement around a point can be modeled in physics as an energy input, which is defined as the metric norm of the local velocity.
     When retrieving your lost ball from the sea, following a straight trajectory wastes energy for nothing.}
    \label{fig:preli_scheme_opti_scheme}
\end{figure}

\paragraph{Distance}
Returning to our objective of learning a measure of distance, we discuss now how such metric gives rise to a Riemannian distance \(\dist_\metric\) on the entire manifold.
First, one can define the length of a smooth path \(\curve\colon [0,1] \to \manifold\) on a Riemannian manifold \((\manifold, \metric)\) in terms of the metric as
\begin{equation}
    \label{eq:length}
    \length(\curve) \coloneqq \int_0^1 \sqrt{\metric_{\curve(t)}(\dot\curve(t),\dot\curve(t))} \dd t\,.
\end{equation}
This path length is invariant to reparametrization of the curve \cite[Lemma 6.1]{lee2006riemannian}.
Furthermore, the path energy is defined as 
\begin{equation}
    \label{eq:path_energy}
    \pathenergy(\curve) \coloneqq \int_0^1 \metric_{\curve(t)}(\dot\curve(t),\dot\curve(t)) \dd t\,,
\end{equation}
which is not independent of the parametrization. 
By the Cauchy-Schwarz inequality, one directly sees
\begin{equation}
    \label{eq:cauchy_energy}
    \length(\curve) \leq \sqrt{\pathenergy(\curve)}
\end{equation}
and equality holds if and only if \(\metric_{\curve(t)}( \dot\curve(t) , \dot\curve(t))\) is constant along the path, which is the case for \textit{geodesics} as we will see later. 
Given the length of paths, we can define the distance between two arbitrary points as the shortest possible length of a path connecting them, which leads to the definition of a \emph{Riemannian distance} \cite[Lemma 6.2]{lee2006riemannian}:
\begin{definition}[Riemannian distance]
    Let \((\manifold, \metric)\) be a Riemannian manifold and \(x,y \in \manifold\) two points on it. 
    Then the \emph{Riemannian distance} between \(x\) and \(y\) is defined as
    \begin{equation}
        \label{eq:dist_opt}
        \dist_\metric(x,y) \coloneqq \inf_{\substack{\gamma\in \mathrm{C}^1([0,1],\manifold)\\\gamma(0)=x, \gamma(1)=y}} \length(\curve).
    \end{equation}
\end{definition}
Shortest paths can have different shapes depending on the metric as illustrated in \Cref{fig:preli_scheme_opti_scheme}.
\begin{example}
    When $\manifold=\SPD^d$ and $\metric$ is the Euclidean scalar product, then $\dist_\metric$ is the usual Frobenius distance.
	When $\metric$ is the affine-invariant metric defined in \Cref{ex:affine_invariant_metric}, the Riemannian distance also has a closed formed expression $\dist_\metric(M_1,M_2) = \left\lVert \log \left(M_1^{-\frac{1}{2}} M_2^{\vphantom{-\frac{1}{2}}} M_1^{-\frac{1}{2}}\right)\right\rVert$ with \(\log\) being the matrix logarithm and \(\lVert \cdot \rVert\) an Euclidean metric on symmetric matrices.
	Note that $\dist_\metric(I,\Sigma)^2=\sum_{i=1}^d \log(\sigma_i)^2 $ with $(\sigma_i)$ the eigenvalues of $\Sigma$ and thus $\dist_\metric(I,\Sigma)^2$ diverges towards the infinity as $\Sigma$ becomes singular.
\end{example}

Reformulating the classical distance metric learning framework with these objects amounts to learning a constant metric $\metric$ defined on a manifold, which corresponds to Euclidean space.
In contrast, \emph{Riemannian} metric learning provides a more general framework, accommodating curved spaces and metrics that vary based on position within the space.
By learning a Riemannian metric, one gains access to additional geometric objects derived from the metric, which we introduce next.
\begin{remark}
    \label{remark:local_approximation}
	Note that not all distances on manifolds are Riemannian distances. 
	For instance, for any injective function $f \colon \manifold\to \mathcal{Y}$, $\dist_f(\cdot,\cdot)\triangleq||f(\cdot)-f(\cdot)||$ is a valid distance on $\manifold$.
	However, a metric $\metric$ such that $\dist_f=\dist_\metric$ does not exist necessarily.
	For example, if \(f=\Id\) maps the unit circle \(\mathsf{S}^1\) to \(\R^2\), the induced distance \(\dist_f\) will approximate the intrinsic distance on the circle only when \(x\) and \(y\) are sufficiently close.
	For distant points on the circle, the map-induced distance will underestimate the true geodesic distance.
	
	However, one can derive a Riemannian metric from a given distance on the manifold by considering its Hessian. 
	The resulting Riemannian metric gives rise to a Riemannian distance that does not necessarily agree with the initially provided one.
	In the example of the circle above, the Hessian would yield to the usual Euclidean scalar product on the tangent space and thus the usual distance on the circle.
	The initially provided distance, however, does yield a local approximation of the Riemannian one and this observation was used by Rumpf and Wirth~\cite{rumpf2015variational} to derive time-discretization of Riemannian calculus.
\end{remark}

In order to find optimality constraints related to \eqref{eq:dist_opt}, we need to compute variations of \(\length\) with respect to \(\curve\) which thus implies second order derivatives $\ddot\curve$.
However, defining second order derivatives via a difference quotient $\lim_{dt\to 0} (\dot\curve(t+dt)-\dot\curve(t))/dt$ is not straightforward on manifolds.
The vectors $\dot\curve(t+dt) \in T_{\curve(t+dt)}\manifold $ and $\dot\curve(t) \in T_{\curve(t)}\manifold $ belong to two different spaces as depicted in \Cref{fig:preli_scheme_tangent_space} and thus cannot we cannot subtract them a priori.
Thus, we introduce a way to do so next.

\paragraph{Covariant Derivative $\nabla$} 
The covariant derivative $\nabla$ is a generalization of the notion of second order derivative and applies to smooth \textit{vector field}.
A smooth vector field \(\vf\) is a map that assigns to every point on the manifold a point in its tangent space, \ie~\(\vf \colon \manifold \to T\manifold\) such that \(\vf(x) \in T_x\manifold\) for any $x\in \manifold$ and which is infinitely differentiable.
The space of smooth vector fields is denoted by $\vecFields$. 
The partial derivatives of the parametrization can be considered as vector fields \(\partial_{i} \chart \colon \manifold \to T\manifold\) that form a basis of the tangent space. 
Hence, any vector field $\vf$ can be written as $\vf(x)=\sum_{i=1}^d V_i(x) \partial_{i}\chart (x)$ for any $x\in \R^d$ where $V_i\colon\manifold\to \R$ are the coordinates in the basis \((\partial_{i})_{i\in[d]}\).
If we aim to differentiate a vector field $\vf$, we will end up differentiating $\partial_{i}\chart$ and thus $\chart$ will be differentiated twice.

We define a covariant derivative $\nabla$ as a map $\vecFields\times \vecFields \to \vecFields $ which satisfies intuitive derivation rules as exposed in the following example or in \Cref{def:covariant_derivative} given in \Cref{sec:appendix_extended_preli} or in \cite[Chapter 4, p.50]{lee2006riemannian}.
Intuitively, \(\nabla_{W}V\) describes how the vector field \(V\) changes when moving in direction \(W\).
In this way, a covariant derivative $\nabla$ of a vector field $\vf=\sum_{i=1}^d \vf_i \partial_i \chart$ in direction of $\wf=\sum_{i=1}^d \wf_i \partial_i \chart$ denoted by $\nabla_\wf \vf$ can be decomposed as 
\begin{align}
    \nabla_\wf \vf&= \sum_{i=1}^d \nabla_\wf (\vf_i \partial_i \chart),\quad (\R-\text{linearity}) \\
    &= \sum_{i=1}^d \nabla_{ \sum_{k=1}^d \wf_k \partial_k \chart} (\vf_i \partial_i \chart) \\
    &= \sum_{i=1}^d \sum_{k=1}^d \wf_k\nabla_{ \partial_k \chart} (\vf_i \partial_i \chart),\quad (C^1-\text{linearity}) \\
    &= \sum_{i=1}^d \sum_{k=1}^d \wf_k\left[\nabla_{ \partial_k \chart} (\vf_i)\partial_i \chart+\vf_i\nabla_{ \partial_k \chart} ( \partial_i \chart) \right],\quad (\text{Product derivation rule})
\end{align}
where $\R-\text{linearity}, C^1-\text{linearity} $ and the product derivation rule can be used because of the definition of $\nabla$. 
Thus, to define a covariant derivative, everything reduce to define $\nabla_{ \partial_k \chart} (\vf_i) $ and $\nabla_{ \partial_k \chart} ( \partial_i \chart)$ for any $k,i\in[d]$.
Since $\vf_i$ is a scalar function and to be consistent with the usual notion of directional derivative, $\nabla_{ \partial_k \chart} (\vf_i)$ is defined as $\nabla \vf_i^\top\partial_k \chart $ where $\nabla \vf_i$ is the usual gradient.
However, there is a priori not a unique way to define $\nabla_{ \partial_k \chart} ( \partial_i \chart)(x)\in T_x\manifold$ for any $k,i\in[d]$ and $x\in \manifold$.
The intuition would be to set $\nabla_{ \partial_k \chart} ( \partial_i \chart)(x)$ as the orthogonal projection of $\partial_k \partial_i \chart(x)$ on $T_x\manifold$ for any $x\in \manifold$.
This is valid only when the metric $\metric_x$ in $T_x\manifold$ is consistent with the Euclidean metric of the ambient space $\R^{d_a}$, \ie~$\metric_x$ is the restriction of the ambient scalar product to $T_x\manifold$.
For more general metric $\metric$, we define the covariant derivative using the basis decomposition for any $i,j\in[d]$,
\begin{equation}
   	\label{eq:param_2nd_deriv}
   	\nabla_{\partial_i \chart} \partial_j\chart = \sum_{k=1}^d \Gamma_{ij}^k \, \partial_{k}\chart 
\end{equation}
where the coordinates are called the \emph{Christoffel symbols} $(\Gamma_{ij}^k \colon \manifold\mapsto \R)_{(i,j,k)\in[d]^3}$ which are smooth scalar functions.
There exists a unique way to define the Christoffel symbols to have symmetry constraints\footnote{Also called torsion-free.} $\Gamma_{ij}^k=\Gamma_{ji}^k$ and a bilinear rule of derivation according to the metric $\metric$, \ie
\begin{equation}
    \label{eq:deriv_metric}
    \nabla_{Z} \metric(\vf,\wf)(x)=\metric_x(\nabla_{Z} \vf(x),\wf(x))+\metric_x( \vf(x),\nabla_Z\wf(x))
\end{equation} 
for any $x\in \manifold$ and $Z,\vf,\wf\in C^\infty(\manifold)$. 
This defines the Levi-Cevita covariant derivative $\nabla$ also called the Levi-Cevita connection \cite[Theorem 5.4]{lee2006riemannian}. 
The following property gives a concrete expression. 
\begin{proposition}
   	\label{prop:christoffel}
   	Let \(\metric^{-1} =  (\metric^{ij})_{(i,j)\in  [d]^2}\) be the inverse of \(\metric\) as before. 
   	Then, the Levi-Cevita covariant derivative has the following representation of the Christoffel symbols:
   	\begin{equation}
   		\label{eq:gammaRepresFinite}
      	\Gamma_{ij}^k = \frac12 \sum_{l=1}^d \metric^{lk} \left(\partial_{i}\metric_{jl} - \partial_{l}\metric_{ij} + \partial_{j}\metric_{li}\right)\, .
   	\end{equation}
\end{proposition}
The Levi-Cevita connection $\nabla$ is particularly suitable since we will see below that locally minimizing curves $\curve$ of the energy \eqref{eq:dist_opt} verify $\nabla_{\dot{\gamma}} \dot{\gamma}=0$ \cite[Corrolary 6.7]{lee2006riemannian}. 
In other words, a local minimizing curve $\gamma$ has a vanishing second order derivative according to the geometry induced by the metric $\metric$. 
See \Cref{sec:appendix_extended_preli} for an infinite dimensional extension of the covariant derivative which is relevant to analyze diffeomorphism groups or Wasserstein space geometry.

\paragraph{Parallel Transport and Geodesics}
With the covariant derivative, we can introduce the notion of parallel vector fields and the related concept of parallel transport \cite[Chapter 4, p.60]{lee2006riemannian}.
\begin{proposition}[Parallel transport]
   	Let \(\gamma \colon \mathsf{I} \rightarrow \manifold\) be a curve.
   	A vector field \(V \colon \mathsf{I} \rightarrow T\manifold\) along \(\gamma\) is called \emph{parallel} if $\nabla_{\dot\curve} V(t) = 0$ for all \(t \in \mathsf{I}\).
   	For \(t_0 \in \mathsf{I}\), \(V_0 \in T_{\gamma(t_0)}\manifold\), there exists a unique parallel vector field \(V \colon \mathsf{I} \rightarrow T\manifold\) with \(V(t_0) = V_0\) following for any $k\in[d]$ the equation
    \begin{equation}
        \label{eq:parallel_transport}
        \dot{V}_k+\sum_{i,j=1}^d\Gamma_{i,j}^k \dot\gamma_i V_j=0.
    \end{equation}
   	Furthermore, the map \(\parTp_{\gamma(t_0) \rightarrow \gamma(t)} \colon T_{\gamma(t_0)}\manifold \rightarrow T_{\gamma(t)}\manifold\), \(V_0 \mapsto V(t)\) is a linear isomorphism.
\end{proposition}

\begin{figure}[t]
    \includegraphics[width=130mm]{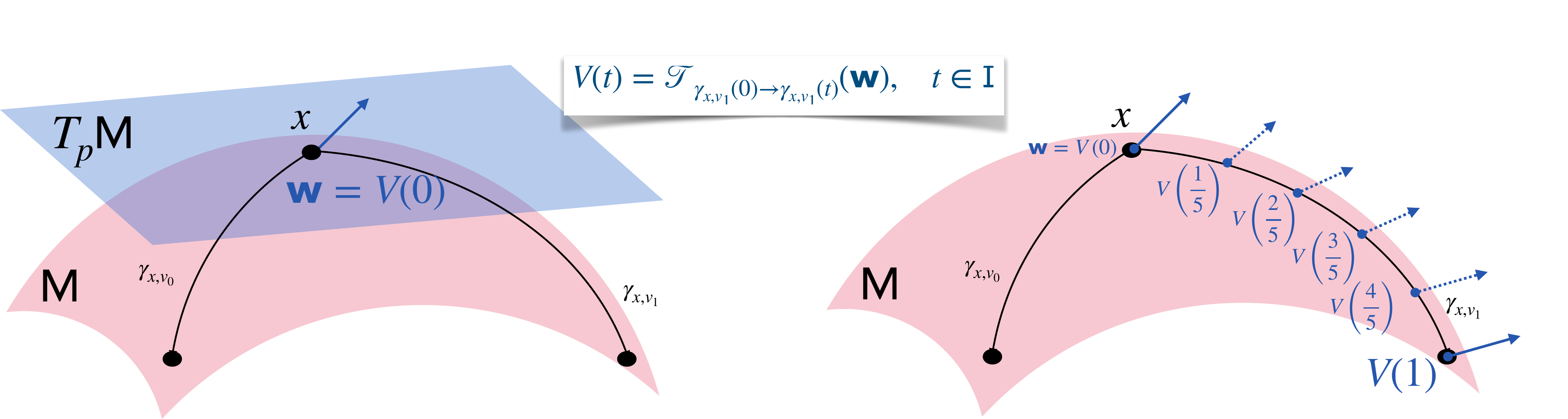}
    \centering
    \caption{Illustration of the parallel transport along the curve $\gamma_{x,v_1}$ (already represented in \Cref{fig:preli_scheme_tangent_space}).}
    \label{fig:preli_scheme_parallel_transport}
\end{figure}
The parallel transport of a vector along a curve is depicted in \Cref{fig:preli_scheme_parallel_transport}.
Finally, the concept of `straight lines' in Euclidean spaces can be generalized to manifolds using \textit{geodesics}.
A geodesic \(\curve\) transports its own velocity vector in a parallel manner \cite[Chapter 4, p.58]{lee2006riemannian}.
\begin{definition}[Geodesic]
    A curve \(\curve \colon \mathsf{I} \to \manifold\) is called \emph{geodesic} if it solves the \emph{geodesic equation}, \ie~the second order ordinary differential equation
    \begin{equation}
        \label{eq:geodesic}
        \nabla_{\dot\curve} \dot\curve(t) = 0 \quad 
    \end{equation}
    for any $t \in \mathsf{I}$.
\end{definition}
Intuitively, this means that a geodesic experiences no acceleration along the direction of the manifold, and its curvature is entirely determined by the curvature of the manifold itself.
This property underscores why geodesics are regarded as generalizations of straight lines.
\begin{example}
    \label{ex:geodesics}
    When using the affine invariant metric on $\SPD^d$, the geodesic $\gamma_{V,\Sigma}$ in direction $V\in \mathcal{S}^d$ with starting point $\Sigma\in \SPD^d$ is $t \mapsto \Sigma^{1/2} \exp(t \Sigma^{-1/2}V\Sigma^{-1/2})\Sigma^{1/2}$. 
    In case of $\Sigma=\Id$, it reduces to the evaluation of the matrix exponential \cite{pennec2006riemannian}.
    The parallel transport of a vector $W\in T_\Sigma \mathcal{S}^d$ at time $s=0$ to time $t$ along the geodesic $\gamma_{V,\Sigma}$ is $\parTp_{\gamma_{V,\Sigma}(0)\rightarrow \gamma_{V,\Sigma}(t)}W=\exp(t V\Sigma^{-1}/2)W\exp(t \Sigma^{-1}V/2)$ \cite[Lemma 3]{schiratti2017bayesian}.
\end{example}

\begin{remark}[Variational definition]
    \label{remark:variational_def}
    A common way to explain geodesics is to describe them as ``shortest paths''. 
    Here, we aim to make this connection between geodesics and distances more precise.
    In fact, every geodesic is a local minimizer of the path length \(\length\).  
    However, it is not necessarily a global minimizer.
    A well-known example of this can be found on a sphere, where geodesics correspond to great circles. For any two points on the sphere, there is always a great circle passing through them.
	The two resulting arcs between the points are both geodesics, but only the shorter of the two arcs represents a global minimizer of the path length, providing the distance between the points.
	Conversely, not every minimizer of the path length is a geodesic, as the curve may not be parametrized by arc length as the path length is invariant under reparametrization.
	Therefore, minimizers of the path length that satisfy the constraint of being parametrized by arc length are indeed geodesics.
	Moreover, minimizers of the path energy $\pathenergy $ are always geodesics.
	Since path energy is not invariant under reparametrization, minimizing the path energy inherently ensures that the path is parametrized by arc length.
	Additionally, as noted earlier, minimizers of the path energy also minimize the path length, reinforcing the connection between the two.
	This has been exploited to define a variational time-discretization often used for shape spaces \cite{hartman2023elastic,heeren2014exploring,rumpf2015variational}.
\end{remark}

\begin{remark}[Hamiltonian flow]
    \label{remark:hamiltonianflow}
     Another way to characterize geodesics is via techniques from geometric mechanics.
     The second-order geodesic ODE \eqref{eq:geodesic} can be reformulated as a Hamiltonian system consisting of two first-order ODEs.
     To achieve this, we introduce the Hamiltonian for any $x,p\in \manifold\times T\manifold$,
    \begin{equation}
        \Hamiltonian(x, \momentum) = \frac{1}{2} \momentum^T \metric_x^{-1} \momentum
    \end{equation}
    and the geodesic equation can be written as the corresponding Hamiltonian flow determined by
    \begin{equation}
        \label{eq:Hamiltonian_equations_easy}
        \frac{d}{dt}x(t) = \frac{\partial}{\partial \momentum}\Hamiltonian(x(t), \momentum(t)), \quad \quad \frac{d}{dt}\momentum(t) = -\frac{\partial}{\partial x}\Hamiltonian(x(t), \momentum(t)).
    \end{equation}
    From this system, the geodesic ODE \eqref{eq:geodesic} is recovered through substitution.
    This approach to the geodesic ODE has been widely used in the shape space literature using the flow of diffeomorphisms \cite{glaunes2006modeling}.
    In our context, it opens up another approach to parametrize the metric structure on the manifold by parametrizing the corresponding Hamiltonian \(\Hamiltonian\). 
    This is related to other problems in physics described via Hamiltonian systems \cite{chen2022learning}.
\end{remark}

Based on any of these definitions, one can introduce the exponential map \cite[Proposition 5.7]{lee2006riemannian}, which ``shoots'' geodesics in prescribed directions as illustrated in \Cref{fig:preli_scheme_tangent_space}.
\begin{definition}[Exponential map]
    \label{def:exponential_map}
    Let \(\curve \colon [0,1] \to \manifold\), be the solution of \(\nabla_{\dot\curve(t)} \dot\curve(t) = 0\) for initial data \(\curve(0)=p\) and \(\dot\curve(0) = V\) if it exists.  
    The Riemannian exponential map \(\Exp_{p} \colon T_p\manifold \rightarrow \manifold\) is defined as \(\Exp_{p}(V) = \curve(1)\).
\end{definition}
This map is called the exponential map because of its link with the exponential on matrices when using the affine-invariant metric on $\SPD^d$ as detailed in \Cref{ex:geodesics}.

\begin{figure}[t]
	\includegraphics[width=130mm]{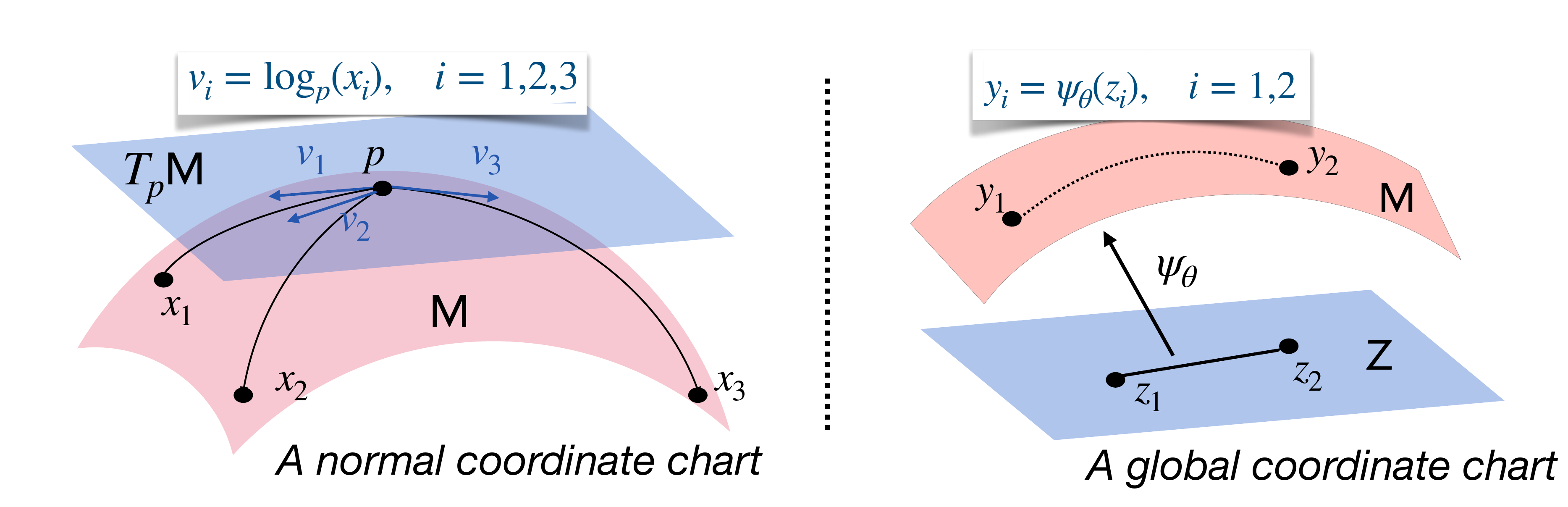}
	\centering
	\caption{Illustration of a normal and a global coordinate chart for feature representations.}
	\label{fig:normal_global}
\end{figure}

In general, one can show that the exponential map is locally bijective, \ie~there exists an \(\eta > 0\), such that \(\Exp_\point \colon B_\eta(0) \to \Exp_\point(B_\eta(0))\) is a bijection denoting by $B_\eta(0)$ the centered open ball of radius $\eta$.
This follows from the local uniqueness of solutions of the geodesic equation.
The image \(U_\point \coloneqq \Exp_\point(B_\eta(0))\) is called a \emph{normal neighborhood} \cite[Lemma 5.10]{lee2006riemannian} of \(\point\) and, in this neighborhood, we can define the Riemannian logarithm as inverse of the exponential map:
\begin{definition}[Logarithm]
    \label{def:logarithm_map}
    The inverse operator of the exponential map is called the Riemannian logarithm \(\log_p \colon U_p \rightarrow T_p\manifold\), where \(U_p\) denotes the normal neighborhood of \(p\). 
\end{definition}
See \Cref{fig:normal_global} for an illustration and \Cref{fig:example_normal} for an application.

\paragraph{Integration}
Finally, to consider probability distributions and statistics on Riemannian manifold, we discuss how to integrate on them \cite[Section II.5]{sakai1996riemannian}.
A metric \(\metric\) introduces a natural volume form $\vol_\metric$ on the manifold \(\manifold\) through its determinant. 
That is, for a smooth function \(f\colon \manifold \to \R\), we obtain the integral
\begin{equation}
    \int_\manifold f(x) \dd \vol_\metric (x) \coloneqq \int_U f(\chart(\coord)) \sqrt{\lvert \det \metric_{\psi(\coord)} \rvert} \dd \coord,
\end{equation}
where the manifold is parametrized using \(\chart \colon U \subset \R^d \to \manifold \).
In case multiple functions are necessary to parametrize \(\manifold\), one defines a partition of unity \cite[Section I.2.1 and p.62]{sakai1996riemannian} corresponding to the cover given by the images of the parametrizations and uses it to define the integral by combining the integrals defined for each image as above.

\subsection{Some Typical Manifolds}
\label{sec:typical_manifold}

Now that we have introduced the structure of Riemannian manifolds, we will take a look at some typical examples of such manifolds and how they come up in the metric learning context.
Progressing from more common and intuitive manifolds to increasingly technical and less familiar ones, we conclude with a general discussion on tractability issues and the benefits of employing such a geometric framework.
Finally, the manifold of Riemannian metrics is introduced as a complex yet promising structure---one that must be tamed for the advancement of Riemannian metric learning.

\paragraph*{Euclidean Space}
Our first example is the most elementary one and, at the same time, by far the most prevalent: the Euclidean space \(\R^d\).
It fulfills all of our requirements for a (Riemannian) manifold using the identity as parametrization.
Nevertheless, one can put a variety of metrics on the Euclidean space, which make it an interesting space to study in metric learning.
Furthermore, it can be used to model of wide range of data, where typical examples in our context include images \cite{annarumma2018deep,chadebec2020geometry,chadebec2022data,tosi2014metrics}, Electroencephalography (EEG) \cite{arvanitidis2016locally}, vector representation of text \cite{zhai2013heterogeneous},  and tabular data such as financial data or sensor measurements in engineering or natural sciences \cite{bansal2024taskmet}.
All the following examples could also be considered in Euclidean space, however, depending on the application we might have additional structure useful to exploit.

\paragraph*{Symmetric Positive Definite Matrices}
When studying manifolds in a differential geometry context, the first examples beyond the Euclidean space are usually curved geometric objects such as spheres or tori.
However, when working with data, matrix manifolds, \ie~sub-manifolds of matrices \(\R^{d\times d}\), are more common and thus our next examples.
The first such is the manifold of symmetric positive definite (SPD) matrices \eqref{eq:SPD}.
It forms an open submanifold of the space of matrices and, more precisely, is an open convex cone, \ie~it is closed under linear interpolation and multiplication with positive scalars.
Symmetric positive definite matrices come up regularly as covariance matrices, \eg~in the analysis of images \cite{jeong2024deep,huang2015log,yger2015supervised,vemulapalli2015riemannian,zhu2018towards} and EEG data \cite{congedo2017riemannian,rozo2025riemann}, as well as for action recognition from motion sensors \cite{harandi2017dimensionality}. 
Furthermore, fields of SPD matrices appear as data in diffusion tensor imaging \cite{pennec2006riemannian,schiratti2017bayesian}.
There are three commonly used Riemannian structures on the SPD manifold: (i) \emph{Euclidean} structures as, for examples, induced by the Frobenius distance \(\dist(M_1, M_2)\coloneqq \lVert M_1 - M_2 \rVert_F\), (ii) \emph{affine-invariant} structures induced by \(\dist(M_1, M_2)\coloneqq \left\lVert \log \left(M_1^{-\frac{1}{2}} M_2^{\vphantom{-\frac{1}{2}}} M_1^{-\frac{1}{2}}\right)\right\rVert\) with \(\log\) being the matrix logarithm and \(\lVert \cdot \rVert\) an Euclidean metric on symmetric matrices, and ultimately (iii) \emph{Log-Euclidean} structures induced by \(\dist(M_1, M_2)\coloneqq \lVert \log M_1 - \log M_2\rVert\) using the same elements \cite{arsigny2007geometric}. 
We repeat here some basic facts about these structures and refer interested reader, for example, to \cite{pennec2006riemannian} for more details.

The Euclidean structures are particularly easy to work with computationally, but they largely disregard the specific structure of SPD matrices, \eg~it is hard to prevent nonpositive eigenvalues using it.
In contrast, affine-invariant structures push symmetric matrices with nonpositive eigenvalues at an infinite distance, but incur high computational costs as it typically leads to an extensive use of matrix inverses, square roots, logarithms, and exponentials.
Log-Euclidean metrics were introduced to achieve a compromise between these two properties. 
Through parametrizing \(\SPD^n\) over the Euclidean space \(\Sym^n\) of symmetric matrices computations become simply Euclidean in the logarithmic domain, limiting the number of necessary evaluations of the matrix logarithm and exponential yet the results are similar to the affine-invariant case \cite{arsigny2007geometric}. 
Furthermore, the Log-Euclidean setting yields a Lie group structure in SPD matrices, \ie~turns it into a space that is both a manifold and a group where multiplication and inversion are smooth mappings. 
We will later discuss in \Cref{sec:parametric} how to parametrize learnable metrics on SPD manifolds using these various structures.

\paragraph*{The Special Orthogonal Group}
Next, we turn to more geometric matrix manifolds. 
The first of which is the \emph{special orthogonal group}
\begin{equation}
	SO(n) \coloneqq \left\{Q \in \R^{n \times n} \mid Q^T Q = \Id_n,\, \det Q = 1 \right\},
\end{equation}
which is a Lie group, \ie~also equipped with a smooth manifold structure.
It is one of the two connected components of the more general \emph{orthogonal group} \(O(n) \coloneqq \left\{Q \in \R^{n \times n} \mid Q^T Q = \Id_n \right\}\) which consists of all distance-preserving linear transformations of \(\R^n\).
The tangent space of \(SO(n)\) (and also \(O(n)\)) at the identity is the Lie algebra of skew-symmetric matrices \(T_\Id SO(n) = \mathfrak{so}(n) \coloneqq \{A \in \R^{n \times n} \mid A^T = -A\}\) and the tangent space at an arbitrary element \(Q\) is given by the transformation of this space under \(Q\), \ie~\(T_Q SO(n) = Q \cdot T_\Id SO(n) \coloneqq \{QA \mid A \in \mathfrak{so}(n) \}\).
Since these tangent spaces are also matrix spaces, to obtain a Riemannian metric one uses the Euclidean metric on matrices and obtains \(\metric_Q(QA, QB) = \metric_\Id(A,B) = \frac{1}{2}\tr A^T B\)\footnote{The factor \(\frac{1}{2}\) is for convenience when defining metrics on the following matrix manifolds.}.
In dimensions two and three, \(SO(n)\) consists of rotations around points or lines respectively and is therefore also called the \emph{rotation group}.
The (special) \emph{Euclidean} group consists of its extension to all distance-preserving affine transformations, \ie~also considering translations: \(SE(n) \coloneqq SO(n) \times \R^n\).
Concerning applications, products of \(SO(n)\) or \(SE(n)\) are commonly used to model the rotation of joints, \eg~of skeleton models or rigs in computer animation \cite{kushnerlearning} or of robotic arms \cite{calinon2020gaussians, beikmohammadi2021learning}.

\paragraph*{The Stiefel Manifold And The Grassmannian}
Our next geometric matrix manifold is the \emph{Stiefel manifold} \cite{edelman1998geometry}
\begin{equation}
	\Stiefel_k(\R^n) \coloneqq \left\{M \in \R^{n \times k} \mid M^T M = \Id_k \right\},
\end{equation}
which consists of orthonormal bases of \(k\)-dimensional subspaces of \(\R^n\). 
It is a compact submanifold of the space of matrices with dimension \(nk-\frac{1}{2}k(k+1)\) and, furthermore, can be constructed also as a quotient manifold by observing \(\Stiefel_k(\R^n) \cong O(n) / O(n-k)\).
Its tangent space is given by 
\begin{equation}
	T_M\Stiefel_k(\R^n) \cong \left\{V \in \R^{n \times k} \mid V^T M = -M^T V \right\}
\end{equation}
and via the quotient manifold construction one obtains the so-called \emph{canonical} metric on it as \(\metric_M(V, W) = \tr V^T (\Id - \frac{1}{2} MM^T) W\).
There is also the \emph{Euclidean} metric on the Stiefel manifold which is obtained by restricting the ambient scalar product on matrices to the tangent space, \ie~\(\metric_M(V, W) = \tr V^T W\).
The relationship between the induced geodesic distances of these two metrics was recently investigated by Mataigne et al.~\cite{mataigne2024bounds}.

In the metric learning context, the Stiefel manifold is typically used to construct the \emph{Grassmannian} \(\Grassmann_k(\R^n)\), which disregards the concrete basis and only considers the linear subspaces, \ie
\begin{equation}
	\Grassmann_k(\R^n) \coloneqq \left\{\mathcal{U} \mid \mathcal{U} \text{ is a } k\text{-dimensional linear subspace of } \R^n \right\}.
\end{equation}
That means it can be described as a further quotient manifold by observing \(\Grassmann_k(\R^n) \cong \Stiefel_k(\R^n) / O(k) \), where the action of \(O(k)\) is by change of the orthonormal basis of the subspace.
We focus here on another description, namely via projection matrices.
For this one identifies a linear subspace with the unique orthogonal projection onto it and obtains
\begin{equation}
	\Grassmann_k(\R^n) \cong \left\{P \in \R^{n\times n} \mid P = P^2 = P^T, \tr P = k  \right\},
\end{equation}
which entails 
\begin{equation}
	T_P\Grassmann_k(\R^n) \cong \left\{V \in \Sym^n \mid V = PV+VP \right\}
\end{equation}
for the tangent space.
Grassmannians occur, for example, in the context of video based face recognition \cite{huang2015projection,hamm2008grassmann,harandi2011graph,huang2014learning}. 
Via the quotient construction, we obtain again a metric that descents from the Stiefel manifold and is simply the Euclidean metric for this description of the tangent space, \ie~\(\metric_P(V, W) = \tr V^T W\).
The corresponding Riemannian distance \(\dist_\metric\) is also called the \emph{Grassmanian distance}.
There is a range of further distances constructed on the Grassmannian, which are not-necessarily Riemannian distances.
The most popular of these distances is the so-called \emph{projection distance} \(\dist(P_1, P_2) \coloneqq \lVert P_1 - P_2 \rVert_F\) which is thus the Euclidean distance of matrices applied to the projections.
It yields a computationally cheap local approximation of the Grassmanian distance, \ie~its differentiation yields the same metric as shown \eg~by Harandi et al.~\cite{harandi2013dictionary}.
This projection distance has been used by a number of approaches to distance/metric learning on Grassmannians such as \cite{hamm2008grassmann,huang2015projection}.
Geometrically, the projection and the Grassmanian distance can also be described using the principal angles between the involved subspaces, which is also true for other proposed distances.
For a list see, for example, \cite[Table 2]{ye2016schubert}.
For a more general introduction into the geometry of the Grassmanian and the Stiefel manifold, see, for example, \cite{bendokat2024grassmann}.

\paragraph*{The Space of Immersions}
The idea of the Grassmannian can be generalized by considering nonlinear, curved submanifolds of \(\R^n\) instead of linear subspaces.
To this end, one considers a base manifold \(M\) with dimension smaller than \(n\). 
Then one studies the space of all its immersions \(\Imm(M, \R^n)\) into \(\R^n\) of sufficient regularity. 
It can be shown that \(\Imm(M, \R^n)\) is indeed a manifold under certain assumptions \cite{hirsch1959immersions}, albeit an infinite-dimensional one\footnote{We do not discuss this here but the ideas from \Cref{sec:preliminaries} can be generalized to infinite-dimensional spaces, see \Cref{sec:appendix_extended_preli}.}. 
When considering closed planar curves, \ie~\(\Imm(S^1, \R^2)\), this manifold can be used, \eg, to model the boundaries of cells in microscopy images. 
Furthermore, surfaces immersed into \(\R^3\) are used in computer graphics to model characters or similar and in medical imaging to model the shape of anatomical objects.
The tangent space of \(\Imm(M, \R^n)\) consists of vector fields along the immersed surface on which various metrics on the space of immersions have been proposed.
Popular choices include Sobolev metric of vector fields \cite{bauer2011sobolev,bauer2024elastic,myers2022regression}, metrics derived from elastic energies \cite{heeren2014exploring}, and finally metrics that descend from metrics on the \emph{space of diffeomorphisms} \cite{arguillere2016diffeomorphic}.
For the latter, one considers the action of diffeomorphisms on immersions by deforming the ambient space and introduces a metric on the Lie group of diffeomorphims of sufficient regularity that is compatible with this action.
Elasticity-derived or Sobolev metrics include parameters weighing different contributions, \eg~different orders of derivatives or different gradient directions (\eg~normal vs.\ tangential).
These parameters are classically derived from physical material properties, however, in the context of Riemannian metric learning, they yield a parametrized family of metrics to optimize over \cite{bauer2024elastic}.

\paragraph*{The Group of Diffeomorphisms}
This concept can also be applied to other spaces on which we have an action of diffeomorphisms.
A very common example for this is the \emph{space of images}, which is usually formalized as square integrable functions on a square domain, \ie~\(L^2([0,1]^n, \R^k)\), where the outputs are color channels or similar.
Here, one considers the left-action of diffeomorphisms (\ie~precomposition with the inverse) and a right-invariant metric on diffeomorphisms (\eg~a suitable Sobolev metric) giving rise to the celebrated LDDMM (Large Deformation Diffeomorphic Metric Mapping) framework \cite{trouve1998diffeomorphisms,varano2017tps,shen2019region,germain2024Shape} which is a popular framework in the medical imaging community and remains an active field of research.

\paragraph*{The Space of Probability Measures}
Another space that is often used to model imaging data is the \emph{space of probability measures}, \ie~the set of all probability measures on a given manifold \(M\).
To introduce a distance (which yields a Riemannian metric) on this space, one turns to \emph{optimal transport}.
Here, one aims to find the most efficient transformation between two probability measures in a certain sense.
Given two measures \(\mu\) and \(\nu\), one aspires to find the joint probability distribution \(\pi\) on the product space \(M \times M\), which has marginals \(\mu\) and \(\nu\) and minimizes \(\int_{M \times M} c(x,y) \dd \pi(x,y)\), where the function \(c \colon M \times M \to \R \) specifies the cost of moving a unit of mass from one place to another.
It is typically derived from the (Riemannian) distance on the base manifold \(\manifold\).
This gives rise to the \emph{\(p\)-Wasserstein distance} on probability measures defined by
\begin{equation}
	\label{eq:Wasserstein}
	\Wasserstein_p(\mu,\nu) \coloneqq \inf_{\pi \in \Pi(\mu,\nu)} \left(\int_{M \times M} \dist_M(x,y)^p \dd \pi(x,y)\right)^{1/p},
\end{equation}
where \(\Pi(\mu,\nu)\) as the set of all couplings as introduced above.
This distance turns the space of probability measures into a formal Riemannian manifold.\footnote{The technical details on the geometry of the space of probability measure are more complicated and go beyond the scope of this review.}
The field of optimal transport originates from studying economic problems and has since been revealed to have deep connections to convexity, partial differential equations, and statistics that have helped advancing several mathematical fields.
More recently, optimal transport has established itself as an important applied tool primarily in mathematical imaging and machine learning.
For a more detailed introduction on this subject which also introduces other formulations and makes many of these connections concrete, we refer to the books by Santambrogio~\cite{santambrogio2015optimal} and by Peyré and Cuturi~\cite{peyre2019computational}.
In the metric learning context, one typically learns a distance on the base manifold \(M\) to obtain a metric on the space of probability measures via the Wasserstein distance.
This has been applied, for example, to learn models for bird migration and single-cell RNA data \cite{scarvelis2023riemannian}, for image classification models \cite{cuturi2014ground,jawanpuria2024riemannian,xu2018multi}, for color transfer \cite{heitz2021ground}, causal inference \cite{farzam2024geometry} and for domain adaptation \cite{kerdoncuff2021metric,zhou2021domain}.
Other metrics have been used on the space of probability measures \cite{tam2023transfer} to address specific application, but they are mostly designed by hand.

\paragraph*{Tractability}

In general, when we leave \(\R^d\) and instead work on one of the above Riemannian manifolds (or others), methods and algorithms become more computationally expensive.
They usually require (optimization) algorithms that are tailored to these manifolds. 
Such algorithms have garnered considerable interest in recent years due to there applicability to a wide range of problems beyond metric learning and there exist standard text books such as \cite{boumal2023introduction}.
For the matrix manifolds introduced above, the computational burden is often more acceptable since the operations needed in optimization---such as the exponential and logarithm---can be evaluated in closed form and considerable research effort has been put into making these efficient.
However, for infinite dimensional spaces, such as spaces of shapes or measures, we often have to first pay the prize of a suitable discretization and then Riemannian operations entail costly numerical algorithms. 
For example, in the case of immersions of surfaces in three-dimensional space one often works with triangle meshes and images are discretized as piece-wise constant functions.

\paragraph*{Geometric Prior}

Nevertheless, considering data to lie on Riemannian manifolds allows us to encode additional structure of the data which is particularly interesting when working in data sparse regimes.
In this case, this additional structure often helps to improve generalization.
For example, van Veldhuizen~\cite{vanveldhuizen2023geometry} showed that using a curved, hyperspherical latent space within Variational Autoencoders improved performances compared to a flat Euclidean one.
Theoretically, Suzuki et al.~\cite{suzuki2021generalization,suzuki2021bgeneralization} proved that embeddings in hyperbolic space offer better statistical generalization guarantees than those in Euclidean space for hierarchical tree-like data.

In the context of metric learning, we of course aim to imprint (some of) the structure of the data onto the metric, not only on the manifold.
By providing a Riemannian metric on an Euclidean space, we also obtain a curved space and, by moving points to infinite distance, we can even encode some topology in the metric \cite{congedo2017riemannian}.
However, in some cases, having the manifold structure already as a strong prior can facilitate the learning process and improve performance, especially concerning generalization.
These two aspects combined yield an intricate decision between how much structure to prescribe explicitly via the manifold and how much to learn via the metric which has to be made on a case-by-case basis and will be illuminated more in the following sections.

\paragraph{The Riemannian Manifold of Riemannian Metrics}
As a mise en abyme, before ending this section, the manifold of Riemannian metrics should not be ommitted \cite{ebin1968space,gilmedrano1991riemannian}.
To avoid technical details, we assume that the base manifold $\manifold\subset \R^d$ is compact and we denote by $\mathcal{G}_\manifold$ the manifold of Riemannian metrics on $\manifold$.
Since Riemannian metric can be seen as a smooth $\SPD^d$-valued function on $\manifold$, the tangent space $T_\metric\mathcal{G}_\manifold $ at $\metric \in \mathcal{G}_\manifold$ is isomorphic to the space of smooth, symmetric, matrix $\mathcal{S}^d$-valued functions on $\manifold$.
The metric on $\mathcal{G}_\manifold$ will thus depend on our choice of metric on $\SPD^d$, where affine-invariant metric is arguably one of the most natural choices \cite{pennec2006riemannian}. 
Then we extend this metric by integration on the manifold\footnote{The integral is finite because the manifold is compact and the integrand is continuous.}, for any $\metric\in \mathcal{G}_\manifold$ and smooth functions $\mathfrak{h},\mathfrak{k} \colon \manifold\to \mathcal{S}^d$
\begin{equation}
	\label{eq:metricofmetric}
	\mathbf{G}_\metric(\mathfrak{h},\mathfrak{k}) \coloneqq \int_{\manifold} \tr(\metric_x^{-1}\mathfrak{h}_x\metric_x^{-1}\mathfrak{k}_x)\dd \vol_\metric(x) 
\end{equation}
Note that the integration relies on the metric volume $\vol_\metric$ in such way that $\mathbf{G}_\metric$ is not a trivial extension of the affine-invariant metric. 
However, the smoothness of Riemannian metrics does not appear explicitly in $\mathbf{G}_\metric$.
Even if working in the manifold $\mathcal{G}_\manifold$ might appear challenging, it turns out that a closed-formed expression is available for the geodesics which is given in \cite[Theorem 3.2]{gilmedrano1991riemannian} even when the base manifold $\manifold$ is not compact.
Nevertheless, the geodesics are expensive to compute.
While this structure seems natural, it is not yet used to perform Riemannian metric learning as we will point out in \Cref{sec:future}.

\section{The Riemannian Metric Learning Problem}
\label{sec:abstract}

We first give this abstract definition of Riemannian metric learning, before detailing its different applications.

\subsection{The Abstract Problem}
\label{sec:abstract_problem}

If a \emph{Riemannian metric learning} method is used, at some point, the dataset $\msx=(x_i)_{i\in [N]}$ has to be generated according to an underlying metric $\metric^*$ related to a manifold $\manifold$, \ie~for any $i\in[N]$, $x_i$ is the realization of $X_i\sim Q_{\metric^*}^i$ 
where $(Q_{\metric^*}^i)_{i\in[N]}$ are probability laws on the input space $\mcx$ and $\manifold\subset \mcx$. 
Note that $X_i$ is not necessarily in $\manifold$ because of possible additive noise $\varepsilon$ such that $X_i=M_i+\varepsilon$ where $M_i\in \manifold$. 
The dataset $\msx$ can have a more complex structure, but we assume that $\msx=(x_i)_i$ here for the sake of simplicity and more complex datasets will be discussed in \Cref{sec:objectives}

Optimizing on the manifold $\mathcal{G}_{\mathcal{M}}$ of all possible infinitely differentiable Riemannian metric on $\manifold$ can be tractable in dimension 1 \cite[Chapter 4-5]{louis2019computational}, but most of times it is not.
Thus, a family of parametrized Riemannian metrics $\mathcal{G}_\theta\subset \mathcal{G}_\manifold$ is chosen. 
The wealth of Riemannian metric learning as an optimization problem arise with the use of different mathematical objects, to name the most commons: the geodesics distance $\dist_\metric$, the exponential map $\Exp_\metric$, the parallel transport $\parTp_\metric$, the metric volume $\vol_\metric$.
Therefore, denoting by $\loss$ the objective function, the abstract form of Riemannian metric learning can be stated as follow :
\begin{equation}
	\label{eq:abstract_problem}
	\argmin_{\metric\in \mathcal{G}_\theta \subset \mathcal{G}_\manifold} \loss(\mathcal{F}(\metric),\msx),\quad  \mathcal{F}(\metric)=(\dist_\metric,\Exp_\metric,\parTp_\metric,\vol_\metric) \eqsp .
\end{equation}
This optimization problem is hard for two reasons:
\begin{enumerate}
	\item The manifold of Riemannian metrics $\mathcal{G}_{\mathcal{M}}$ has a specific structure \cite{gilmedrano1991riemannian} (see \eqref{eq:metricofmetric}).
	They are not only vectors or constrained-matrices in $\SPD$, they are $\SPD$-valued smooth functions with a manifold input space $\manifold$. $\mathcal{G}_{\mathcal{M}}$ is both constrained and infinite dimensional.
	This makes it challenging to find suitable discretizations of it, which will be further discussed in \Cref{sec:parametric}. 
	\item The mathematical objects depending on the metric $\mathcal{F}(\metric)$ have, in general, not a closed form formula and their dependence on $\metric$ is not linear. 
	Thus, the optimization problem is not convex except when restricting to constant Riemannian metrics.
\end{enumerate}
However, \emph{Riemannian metric learning} not only offers inescapable problems, but also yield new opportunities for representation learning to leverage a large number of data in an unsupervised manner as we will see in \Cref{sec:how}.
Before we \emph{how} to learn a Riemannian, we will explain \emph{why} by looking at the applications in the next section.

\subsection{Applications: Why Learn a Metric?}
\label{sec:applications}


First, we will present the applications related to the \emph{distance metric learning} literature (\Cref{sec:distance_metric_learning}), then the applications related to \textit{Riemannian metric learning} (\Cref{sec:applications_riemannian_metric_learning}).
When the learned distance $\dist_\metric$ is induced by a non-linear Riemannian metric $\metric$, the application is presented in \Cref{sec:applications_riemannian_metric_learning} instead of \Cref{sec:distance_metric_learning}.
We aim to showcase the richness of the Riemannian metric learning approach and to highlight the importance of each Riemannian object of interest, namely $(\dist_\metric, \Exp_\metric, \parTp_\metric, \vol_\metric)$.
Given that Riemannian metric learning is inherently technical and complex, it is crucial to ensure its relevance in the context of the application. 

\subsubsection{Distance Metric Learning}
\label{sec:distance_metric_learning}
Learning the metric of a metric space (see \Cref{def:metric_space}) can be seen as a fondational task which has its most prominent use in distance-based regression, classification (k-NN), clustering and dimension reduction.
Instead of specifying a distance according to your own prior, you can learn relevant information from the dataset and introduce it in your distance.

\paragraph*{Regression, Classification and Clustering}

On Euclidean data, the paper \cite{xing2002distance} popularized metric learning for clustering and classification using nine dataset from the UC Irvine repository\footnote{https://archive.ics.uci.edu} (UCI) by extending k-means \cite{ahmed2020k}.
This classical repository is composed principally of tabular dataset having less than 3000 samples and 10 classes with low dimensional inputs ($d$<100).
Then, it was followed by \cite{shalev2004online,goldberger2004neighbourhood,globerson2005metric,davis2007information,weinberger2009distance,kedem2012non,wang2014kernel,zadeh2016geometric,chen2018adversarial} outperforming each other chronologically regarding accuracy and time of computations on UCI datasets, Handwritten text recognition (MNIST\footnote{https://yann.lecun.com/exdb/mnist/}, USPS\footnote{https://paperswithcode.com/dataset/usps}) and Face recognition datasets such as Yales \cite{belhumeur1997eigenfaces}, PubFig \cite{nair2010rectified}, LFW \cite{huang2008labeled}. 

A unified benchmark was performed in the review \cite{suarez2021tutorial} and all methods are implemented in the related library pydml \cite{suarez2020pydml}.
Interestingly, one of the last paper \cite{chen2019curvilinear} improving on these methods suggest to learn a curved coordinate system to perform metric learning, which is close to Riemannian metric learning.
This method presented was then used successfully for recommendation systems in \cite{zhang2021curvilinear}.

These methods were then extended to other types of data such as multimodal data \cite{mcfee2011learning,xie2013multi,zhai2013heterogeneous}, sets \cite{zhu2013point}, bag of instances \cite{guillaumin2010multiple} and to other contexts such as Multitask learning \cite{xu2018multi,guillaumin2010multiple,parameswaran2010large}. 
The applications were face, object, and artist recognition in images, videos, and audio. 
Using metric learning within multitask kNN classifiers \cite{guillaumin2010multiple} is particularly appealing since the number of parameters does not depend on the number of classes contrary to multitask-SVM \cite{evgeniou2004regularized}.

Then, a new line of works arose in computer vision.
Covariance matrices and subspaces features arising from images and videos were embedded in the Grassmanian \cite{hamm2008grassmann,harandi2011graph,huang2015projection,huang2014learning,zhu2018towards} and the SPD manifold \cite{harandi2017dimensionality,huang2015log,yger2015supervised,vemulapalli2015riemannian,zhu2018towards} and distances on these non-trivial manifolds were learned. 
Benchmarks were performed on objects recognition task using the ETH-80 datasets \cite{leibe2003analyzing}, on material recognition using the Flickr material dataset \cite{sharan2009material}, UIUC material dataset \cite{liao2013non}, and on face recognition using the Youtube celebrities \cite{kim2008face} and Youtube Face dataset \cite{wolf2011face}. 
One of the latest papers in this direction, \cite{zhu2018towards}, offered a general method working both on the SPD and Grassmanian manifold and further improved accuracy and computation time.

As mentioned in \cite{kaya2019deep,roth2020revisiting}, deep metric learning methods have widened the scope of applications and improved on the state-of-the-art on different task of computer vision \cite{mohan2023deep}: visual tracking \cite{hu2015deep}, face recognition \cite{chen2019distance}, person re-identification \cite{hermans2017defense}, labeling radiography \cite{annarumma2018deep}, 3D objects retrieval \cite{he2018triplet,bronstein2010data}, semantic instance segmentation \cite{fathi2017semantic}, semantic edge detection \cite{cai2017semantic}, pan-sharpening \cite{xing2018pan}, volumetric image recognition \cite{wang2018multi}, crowdedness regression \cite{wang2017deep}.
The methods apply also in audio and natural language processing with topic classification \cite{ein2018learning}, music ranking \cite{lu2017deep}, and speaker recognition \cite{wang2019centroid} using so-called prototypical networks \cite{snell2017prototypical}.
However, the success of deep metric learning methods was debated by Musgrave et al.~\cite{musgrave2020metric}, who raise issues about the lack of clarity and fairness in the experiments.

\paragraph*{Dimension Reduction}
\label{para:dim_red}
The task of dimension reduction \cite{wang2015survey,meilua2024manifold}, also called \textit{manifold learning}, is used to visualize high dimensional datasets or to pre-process data by removing unnecessary components.
Most methods such as Multidimensional Scaling (MDS) \cite{cox2000multidimensional} or Stochastic neighborhood embedding (SNE) \cite{hinton2002stochastic} reduce the dimension while preserving global or local distances from the raw dataset.
That is why, manifold learning methods can benefit from learning a distance.

Moreover, if the distance metric learning method uses a projection matrix $L$, such as in \cite{shalev2004online,goldberger2004neighbourhood,globerson2005metric,davis2007information,weinberger2009distance,kedem2012non,wang2014kernel,chen2018adversarial}, the eigenvectors related to the largest eigenvalues of $L$ can also be used directly to project the dataset in small dimensions as explained in the review dedicated to this topic \cite{wang2015survey}.

There are also methods employing \emph{Riemannian metric learning} for dimension reduction which are given in the next sub-section.

\subsubsection{Riemannian Metric Learning}
\label{sec:applications_riemannian_metric_learning}

As discussed in \Cref{sec:abstract_problem}, a Riemannian metric can be used for modeling through different objects: the distance $\dist_\metric$, the exponential map $\Exp_\metric$ related to geodesics $\curve$, the parallel transport $\parTp_\metric$, the metric volume $\vol_\metric$ or even sometimes the Ricci curvature of the manifold $R_\metric$ \cite{li2023geometry,farzam2024geometry}.
In the following, we clarify and exemplify why these objects can be useful for representation learning, Hamiltonian learning, temporal longitudinal modeling, distribution trajectory inference, sampling and generative modeling, domain adaptation, and finally causal inference.

\paragraph*{Representation Learning}
The first paper we found with ``Riemannian metric learning'' in its title was by Peltonen et al.~\cite{peltonen2004improved} in 2004 for exploratory analysis and dimension reduction.
They aruged that, although Riemannian metric learning increased the computational burden, they used it since global distance would not have preserved the topology of the data space, which they deemed important for identifying subcategories of the classes within a dataset \cite[p.3]{peltonen2004improved}. 
Other papers also used principles of Riemannian metric learning, without saying it, for instance Isomap \cite[2000]{tenenbaum2000global} is arguably one of the most famous.
In Isomap \cite{tenenbaum2000global}, the geodesic distance $\dist_\metric$ is defined according to the k-nearest neighbors graph, which can cause robustness issues \cite{balasubramanian2002isomap}.
Recently, a variant of Isomap using parallel transport $\parTp$ proposed by Budninskiy et al.~\cite{budninskiy2019parallel} revealed to be more robust. 
Therefore, representations can be learned through Riemannian metric learning, not only by using distances but also by using connections on tangent bundles.

\begin{figure}[t]
	\includegraphics[width=130mm]{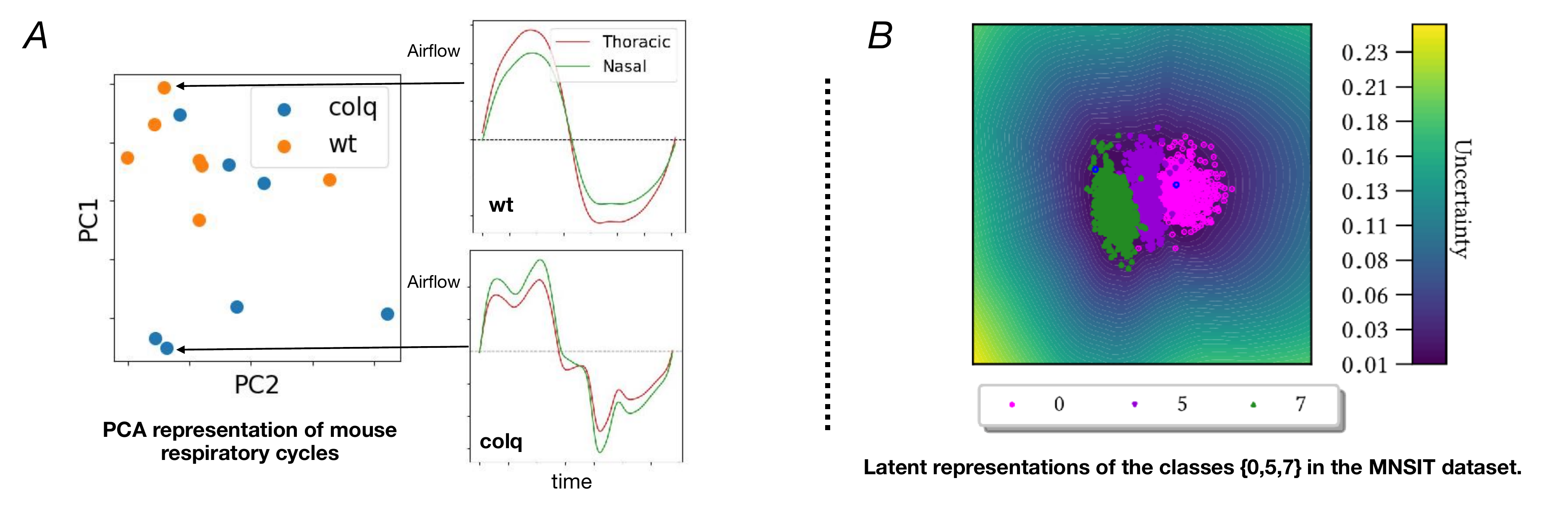}
	\centering
	\caption{ 
		On the left (A), an example taken from \cite{germain2024Shape} (with permission) of normal coordinate chart representing shapes of respiratory cycles related to two mouse genotypes: colq and wt.
		On the right (B), an example taken from \cite{syrota2024decoder} (licensed under \href{https://creativecommons.org/licenses/by/4.0/}{CC BY 4.0}) of latent representation where the metric encodes uncertainty information related to an ensemble of decoders $\psi_\theta$.
	}
	\label{fig:example_normal}
\end{figure}

Another line of work, learn representations via \emph{normal coordinate charts} \cite{lin2008riemannian} using the logarithm and exponential map.
If a general metric $\metric$ is learned, a point cloud $(x_i)_{i\in[N]}$ on the related manifold $\manifold$ can be represented by their logarithms $(\log_p(x_i))_{i\in[N]}$ taken at their Fréchet mean $p$ defined as 
\begin{equation}
	\label{eq:karcher_mean}
	p=\argmin_{q\in \manifold} \sum_{i=1}^N \dist_\metric(q,x_i)^2\quad .
\end{equation}
Then, any standard method of dimension reduction (\eg~PCA) can be applied to these representations $(\log_p(x_i))_{i\in[N]}$ \cite{hauberg2012geometric,lin2008riemannian,varano2017tps,germain2024Shape}.
This strategy can thus transform points $(x_i)_i$ on a Riemannian manifold $\manifold$, where there is not necessarily a group structure, to vectors $(\log_p(x_i))_{i\in[N]}$ in a vector space $T_pM$, where there is a group structure.
This opens the toolbox of Euclidean statistical methods, which is appealing to analyze complex objects such as \emph{shapes} \cite{vaillant2004statistics} as depicted in \Cref{fig:example_normal} with mouse respiratory cycles.

Simpler than a normal coordinate chart is a global coordinate chart $\psi$ as used in \Cref{sec:preliminaries} and implicitly in autoencoders \cite{bank2023autoencoders} and variational autoencoders \cite{kingma2019introduction}.
Here, the observations $y=\psi_\theta(z)$ are represented by low dimensional latent variables $z\in \mathcal{Z}$.
This idea was further developed in \cite{kalatzis2020variational,frenzel2019latent,chadebec2020geometry,arvanitidis2021geometrically,fasina2023neural,diepeveen2024pulling,diepeveen2024score}
to enhance training and interpretability of the representation through visualization of geometric features, which might also encode task-related uncertainty \cite{warburg2024bayesian,syrota2024decoder} as depicted in \Cref{fig:example_normal}.
Once a latent space and its geometric structure $(\mathcal{Z},\metric)$ are learned, different applications involving Riemannian distance, geodesics, or simply features representation can be performed.
Estimates of manifold curvature are even employed in the study of brain neural activation signals \cite{acosta2023quantifying}.
Geodesics in the latent space are especially appealing for temporal longitudinal modeling and distribution trajectory inference as detailed in the next sections.

\begin{figure}[t]
	\includegraphics[width=130mm]{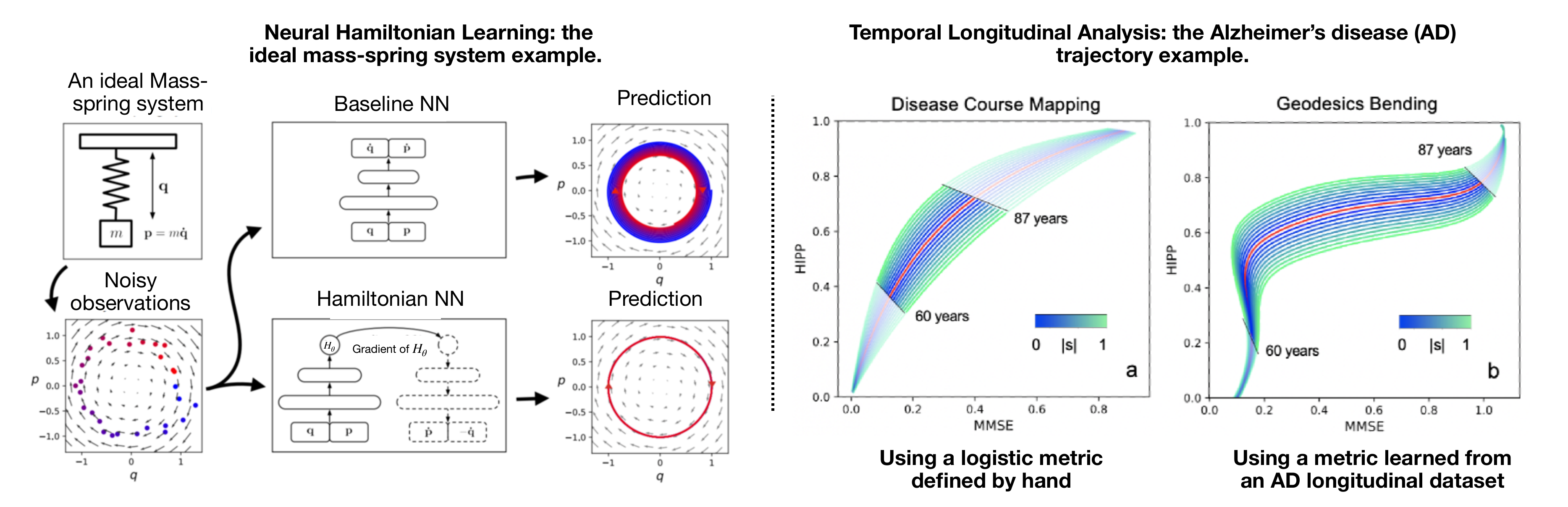}
	\centering
	\caption{
		On the left, an example of Hamiltonian learning taken from \cite{chen2022learning} (licensed under \href{https://creativecommons.org/licenses/by/4.0/}{CC BY 4.0}).
		Neural Hamiltonian learning is compared to a naive neural networks encoding the system by its differential equation. 
		Note that predicted baseline trajecories are thicker because of the larger uncertainty of estimation on the derivatives compared to the Hamiltonian.
		On the right, an example taken from \cite{gruffaz2021learning} (with permission) where the geodesics are learned to mimic AD patient trajectories. 
		HIPP is the cerebral volume of the Hippocampus and MMSE refers to the Mini Mental Status Examination score. 
		The scores were reversed and normalized in order to be increasing in $[0,1]$. 
		The learned exponential shape in figure b is consistent with clinical observations, but is not present when considering a hand-cafted logistic metric \cite{schiratti2017bayesian} showing the interest of learning the Riemannian metric from scratch.
		The tubular green and blue curves are the exp-paralellization of the average geodesic in red for different values of space shifts $w$ \eqref{eq:exp_para}, it represents the inter-indiviudal variability of trajectories along the mean trajectory.
	}
	\label{fig:trajecory_example}
\end{figure}

\paragraph*{Hamiltonian Learning}
Before presenting the applications related to geodesics modeling, we include a discussion on the broader context of Hamiltonian learning. 
In physics, trajectories are often analyzed using Hamiltonians, which describe preserved energy along the trajectory.
The Hamiltonian related to an object at position $q$ and momentum $p$ is denoted by $\mathcal{H}(q,p)$ and decomposes as $\mathcal{H}(q,p)=V(q)+p^\top \metric(q)^{-1} p/2$ with $V(q)$ being the potential energy at position $q$ and $p^\top \metric(q)^{-1} p/2$ the kinetic energy depending on the metric of the space. 
Recently, deep learning methods were suggested \cite{toth2019hamiltonian,chen2022learning} to model the underlying Hamiltonian as a neural network and infer it from observed trajectories. 
This is depicted on the left of \Cref{fig:trajecory_example}.

Neural Hamiltonians connect with Riemannian metric learning since geodesics are Hamiltonian trajectories for the special case of \(V\equiv0\) as mentioned in \Cref{remark:hamiltonianflow}.
When the kinectic energy is fixed within Neural Hamiltonians, \ie~$\metric$ constant, we only need to analyze the potential $V$ to get insight about the state space, which can enhance the interpretability of the Hamiltonian learning \cite{souveton2024fixed}.
In the other way around, if $V$ is removed or fixed, learning the metric $\metric$ shed light not only on states $q$, but also on displacement $p$.
It is not only about where your are, but where you want to go.
In the next paragraph, we illustrate these ideas with Longitudinal modeling and Trajectory inference.

\paragraph{Temporal Longitudinal Modeling}
In various scientific fields, a key objective is to deduce the dynamics of an underlying system from noisy data.
This is particularly challenging in areas such as biomedicine, where only limited longitudinal data is available---such as when tracking health indicators or studying disease progression \cite{hay2021estimating}.
For example, in Alzheimer's disease follow-up, pyschometric assessment and brain imaging are available at different times depending on the patient \cite{marinescu2018tadpole}.

In temporal longitudinal modeling\footnote{This is not the general case of longitudinal modeling. In longitudinal modeling the outcome variable $Y(x)$ might depends on a covariate $x$ which is not a scalar as timepoints.},
the same individual $i$ is observed with features $(y_i^j)$ at different time points $(t_i^j)$ which can be irregularly sampled.
The goal is to recover the underlying process $t\to Y_i(t)$ such that $Y_i(t_i^j)=y_i^j+\varepsilon_i^j$ with $\varepsilon_i^j$ an \iid~noise.
One approach is to model the individual trajectories $Y_i$ as geodesics, \ie~for any $t\in[0,1]$, $Y_i(t)=\Exp_{q_i}(tv_i)$ with $(q_i,v_i)\in \manifold \times T\manifold$.
The idea is that geodesics generalized straight lines which make them both flexible and simple enough to avoid overfitting on noisy datasets.
Geodesics learning was also used to model trajectories on rotations manifold to interpolate animations \cite{kushnerlearning}, human or robot motion captures \cite{beikmohammadi2021learning,tosi2014metrics}, image registration \cite{shen2019region,myers2022regression}, and taxi routes in traffic jams \cite{benmansour2010derivatives,qiu2024estimating}.
We should note that geodesic learning can also occur in exotic application such as a two player game where the goal of the first player is to install a surveillance system and the goal of the second is to reach a target area without being detected \cite{mirebeau2017automatic}.

In the following, we detail the case of Alzheimer's disease (AD) follow-up to enlighten how Riemannian metric learning can be relevant in a specific application.
Geodesics are particularly suited to model AD follow-up \cite{lorenzi2015disentangling,schiratti2017bayesian} since AD is a neurodegenerative disease, \ie~there is rarely recovery of patients, which implies that there are no loops in patient trajectories \cite{jeong2024deep}. 
Moreover, there is an inter-patient variability in the AD trajectories.
That means two AD patients can have a similar trajectory direction because of their disease, but their trajectories remain parallel and distinct because of their individual health conditions.
This variability can be identified within mixed-effect models by generalizing the concept of parallel straight lines using the exp-parallelization \cite{schiratti2017bayesian}.
The exp-parallelization $\operatorname*{ExPar}(w)(\curve,\cdot)$ of a curve $\curve \colon \mathsf{I}\subset \R\to \manifold $ at $t_0\in\mathsf{I}$ in the direction $w\in T_{\curve(t_0)}\manifold$ is defined as,
\begin{equation}
	\label{eq:exp_para}
	\operatorname*{ExPar}(w)(\curve,t)=\Exp_{\curve(t)}(\parTp_{\curve,t_0,t}(w))\eqsp,
\end{equation}
for any $t\in \mathsf{I}$.
This shows how parallel transport and the exponential can be combined when optimizing a Riemannian metric.
Note that the exp-parallelization of a geodesic is not necessarily a geodesic.
An illustration of the transformation is given on the right of \Cref{fig:trajecory_example}.
Alzheimer's disease does not start at the same age for everyone and does not evolve at the same speed.
This generates inter-patient temporal variability.
Thus, it is crucial to compose an exp-parallelization with an individual time warp $\psi_i\colon\mathsf{I}\mapsto \R$ as $Y_i=\operatorname*{ExPar}(w_i)(\Exp_{q_0}(v_0\cdot),\psi_i)$ with $(q_0,v_0,w_i)\in \manifold \times T_{q_0}\manifold  \times T_{q_0}\manifold$.
Parameters related to time ($\psi_i$) and space ($q_0,v_0,w_i$) are statisticaly identifiable as long as $||v_0||=1$ since the geodesics are parametrized by arc length (\Cref{remark:variational_def}).
This property is particularly interesting to disentangle time and space variability in the analysis. 

This line of works was thus naturally extended by Riemannian metric learning methods \cite{louis2019computational,louis2019riemannian,gruffaz2021learning,sauty2022riemannian}.
Here, the goal was to learn the best metric related to the mixed-effect model to represent the patients' trajectories.
The resulting representation of the patient trajectories can then be analyzed to understand the disease progression according to different health conditions.
This proved successful to handle multimodality \cite{sauty2023multimodal,gruffaz2021learning,fournier2023multimodal}.

\paragraph*{Distribution Trajectory Inference}
In single-cell RNA sequencing (scRNA), cross-sectional data collection is sparse and static due to the high costs and destructive nature of the procedure \cite{macosko2015highly}.
As a result, there is a need to create models capable of reconstructing a system's temporal dynamics (for example, in cellular studies) from observed samples taken at discrete time points.
This general problem is known as \emph{trajectory inference} \cite{lavenant2024toward}.
In trajectory inference, we only access the population distribution $(P^j)_{j\in[T]}$ at different fixed time points $(t^j)_{j\in[T]}$.
The goal is to recover the continuous trajectory of the distribution by interpolation.

\begin{figure}[t]
	\includegraphics[width=130mm]{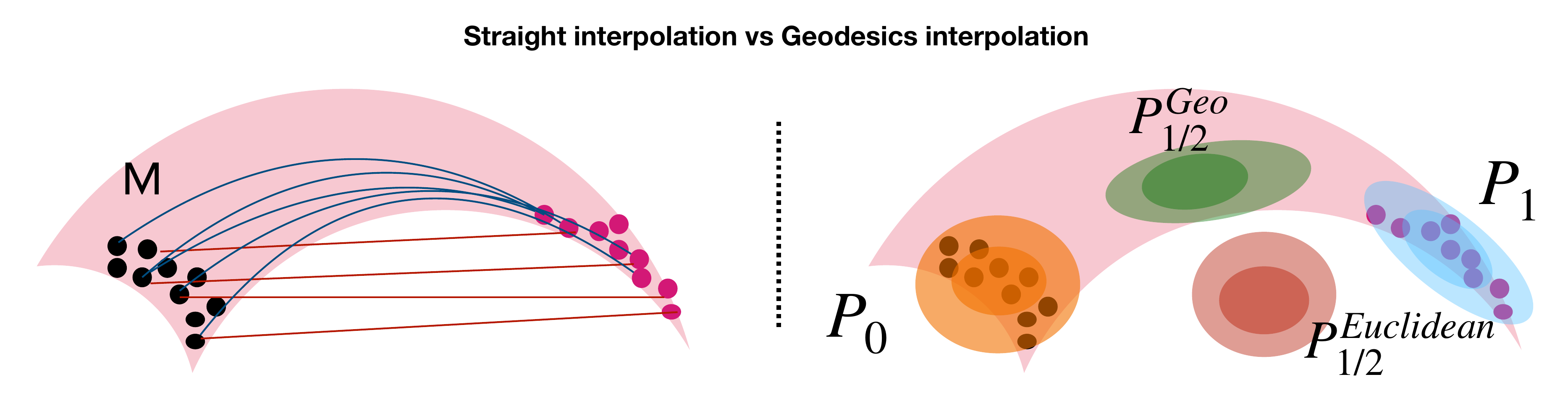}
	\centering
	\caption{
		Example of trajectory inference with different choices of interpolant. 
		$P_{1/2}^{Euclidean}$ and $P_{1/2}^{Geo}$ are the marginal density at time $t=1/2$ when the interpolant uses respectively straight lines or geodesics, respectively. 
	}
	\label{fig:flow_matching}
\end{figure}

One can approach this problem via optimal transport \cite{santambrogio2015optimal}:
For any $j\in[T-1]$, a transport map $T^j$ is learned such that $P^{j+1}=T^j\#P^j$ is the pushforward measure of $P^j$ by $T^j$. 
When using the Euclidean metric as base metric, the interpolation becomes $P^{j+t}=(tT^j+(1-t)\Id)\#P^j$ for any $t\in[0,1]$.
However, this Euclidean interpolation can put some mass in regions of the space which are unrealistic. 
A conceptual illustration is given in \Cref{fig:flow_matching}.
That is why Scarvelis and Solomon~\cite{scarvelis2023riemannian} and Huguet et al.~\cite{huguet2022manifold} proposed to learn a Riemannian metric to be used in the optimal transport problem. 
In the end, for any $x$ in the observational space, defining $\gamma(x,t)$ as the geodesics between $x$ and $T^j(x)$ for any $j\in[T-1]$, the interpolation becomes $P^{j+t}=\gamma(\cdot,t)\#P^j$. 
The method improved the results on a bird migration dataset\footnote{https://birdsoftheworld.org/bow/species/snogoo/cur/movement} and an scRNA dataset \cite{schiebinger2019optimal}.

Flow matching \cite{lipman2023flow} is another option, where the interpolation of distributions is achieved by integrating a flow equation $\frac{\dd X}{\dd t}=v_t(X)$. 
Here, the velocity field is learned in order to match a quantity related to an interpolating function $I\colon\manifold^2\times [0,1]\to \manifold$.
This function is a prior on the shape of trajectory we want after integrating the flow equation.
Most of the time, $I$ is chosen to be linear because of its simplicity.
However, as for Euclidean optimal transport interpolation, this choice is not optimal since the interpolation might leave the data manifold as depicted in \Cref{fig:flow_matching}.
Again, a Riemannian metric can be learned from the dataset and geodesics can be used to define the desired interpolant as $I(x,y,t)\approx\Exp_x(t\Log_x(y))$ using neural network approximations \cite{kapusniak2024metric,sun2025geometry}.
Kapusniak et al.~\cite{kapusniak2024metric} and Sun et al.~\cite{sun2025geometry} successfully used Riemannian metric learning within flow matching for single-cell trajectory inference on the CITE-seq and Multiome datasets from a NeurIPS competition \cite{burkhardt2022multimodal}, where both methods greatly improved on the state-of-the-art.

\begin{figure}[t]
	\includegraphics[width=130mm]{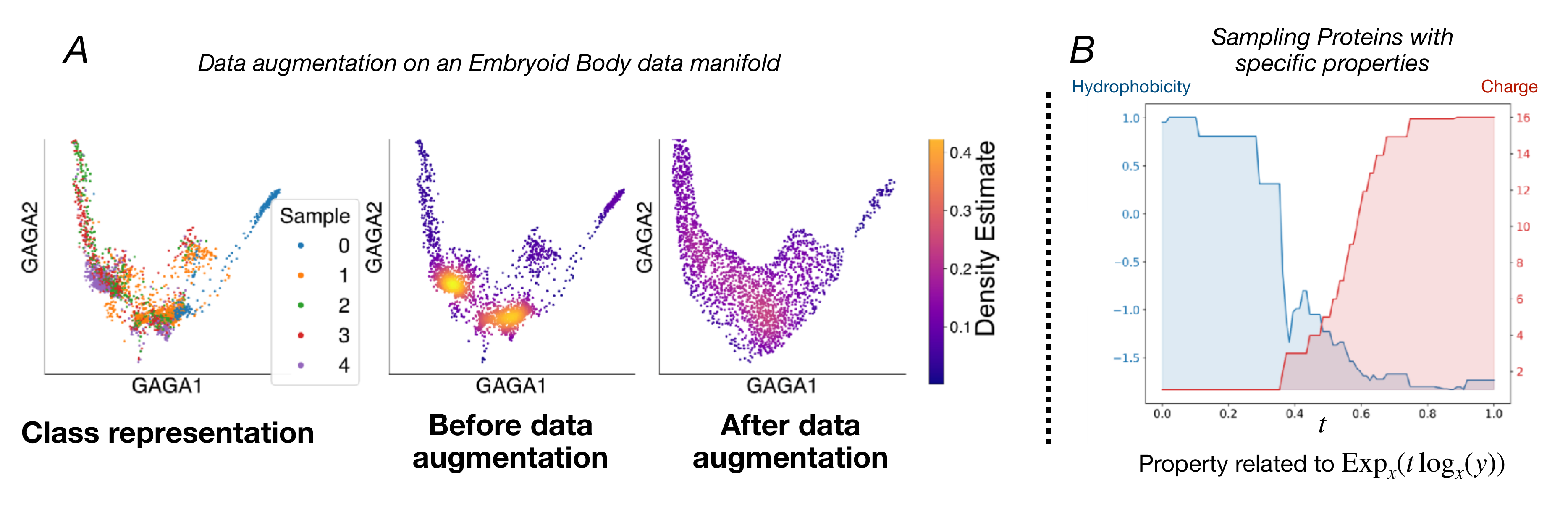}
	\centering
	\caption{
		On the left, a Riemannian manifold strcuture is learned \cite{sun2025geometry} (licensed under \href{https://creativecommons.org/licenses/by/4.0/}{CC BY 4.0})  using embryoid data and data augmentation is performed by sampling according to the metric volume.
		On the right, synthetic proteins are generated in \cite{dekruiff2024pullback} (licensed under \href{https://creativecommons.org/licenses/by/4.0/}{CC BY 4.0})  by interpolating a protein with high hydrophobic moment and low charge $x$ and an another one with a low hydrophobic moment and high charge $y$.
	}
	\label{fig:sampling_example}
\end{figure}

\paragraph*{Sampling and Generative Modeling}
More generally, Riemannian metric learning finds also applications in sampling \cite{cui2024optimal,chadebec2020geometry,chadebec2022data} and generative modeling \cite{arvanitidis2016locally,frenzel2019latent,arvanitidis2021geometrically}.
In these applications, the metric $\metric$ and its volume $\vol_\metric$ are particularly involved. 
In \cite{girolami2011riemann}, the authors demonstrate that sampling algorithms such as Hamiltonian Monte Carlo (HMC) \cite{betancourt2017conceptual} can be improved by using a Riemannian metric related to the distribution.
Chadebec et al.~\cite{chadebec2020geometry} built on this idea by learning the metric on a latent space in a variational autoencoders to perform data augmentation for brain MRI \cite{chadebec2022data}.
Learning more structure in the latent space is particularly interesting \cite{chadebec2022data,vanveldhuizen2023geometry} when datasets are small and high dimensional---such as MRI datasets. 
Sun et al.~\cite{sun2025geometry} suggest to sample uniformly according to the volume $\vol_\metric$ in the Riemannian latent space in order to avoid selection bias due to observations using the unadajusted Langevin algorithm as depicted on the left of \Cref{fig:sampling_example}. 
Cui et al.~\cite{cui2024optimal} remind us that the convergence guarantee of Langevin dynamics to their invariant distribution is related to Poincaré inequalities. 
Thus, they optimize the Poincaré constant to accelerate convergence by learning a Riemannian metric on the state space.
Therefore, Riemannian metric learning can be a tool to improve sampling efficiency motivated by convergence results.

In an unsupervised learning context, Arvanitidis et al.~\cite{arvanitidis2016locally} suggested to learn a Gaussian mixture on a Riemannian manifold adapted to a sleep stages dataset from Physionet.
They adapted the geometry of the Gaussian in order to avoid holes in the data distribution.
Using autoencoders and variational autoencoders, several papers \cite{arvanitidis2018latent,frenzel2019latent,arvanitidis2022prior,arvanitidis2021geometrically,dekruiff2024pullback,rozo2025riemann} have underlined that learning the Riemannian structure of the latent space improves uncertainty estimation and interpretability.

For instance, de Kruiff et al.~\cite{dekruiff2024pullback} they learn an isometric map between the latent space and the observational space in order to preserve distance related to biological properties of observed proteins.
By using the Riemannian structure of their latent manifold, they can generate new proteins with specific biological properties with simple geodesic interpolation \cite[Figure 5]{dekruiff2024pullback} as depicted on the right of \Cref{fig:sampling_example}.  
A score-based variant of the method is given in \cite{diepeveen2024score} and its theoretical underpinnings are discussed in more detail in \cite{diepeveen2024pulling}.

\paragraph*{Domain Adaptation and Transfer Learning}
The main goal of domain adaptation and transfer learning is the following to generalize a learned statistical model from a source distribution $P_S$ to a target distribution $P_T$ using weak information.
For instance, in the case of classification, the target features are available but not their labels \cite{baktashmotlagh2013unsupervised}.

Recently, optimal transport has become a key tool in domain adaptation. 
For classification, the target features are transported on source features by minimizing the global cost of displacement and then the statistical model estimated on the source domain can be applied.
The cost of displacement, \ie~the ground metric, can be learned to improve performance \cite{buttazzo2004optimal,cuturi2014ground,heitz2021ground,jawanpuria2024riemannian}.
This way domain adaptation benefits from distance metric learning \cite{kerdoncuff2021metric,zhou2021domain}.

\begin{figure}[t]
	\includegraphics[width=130mm]{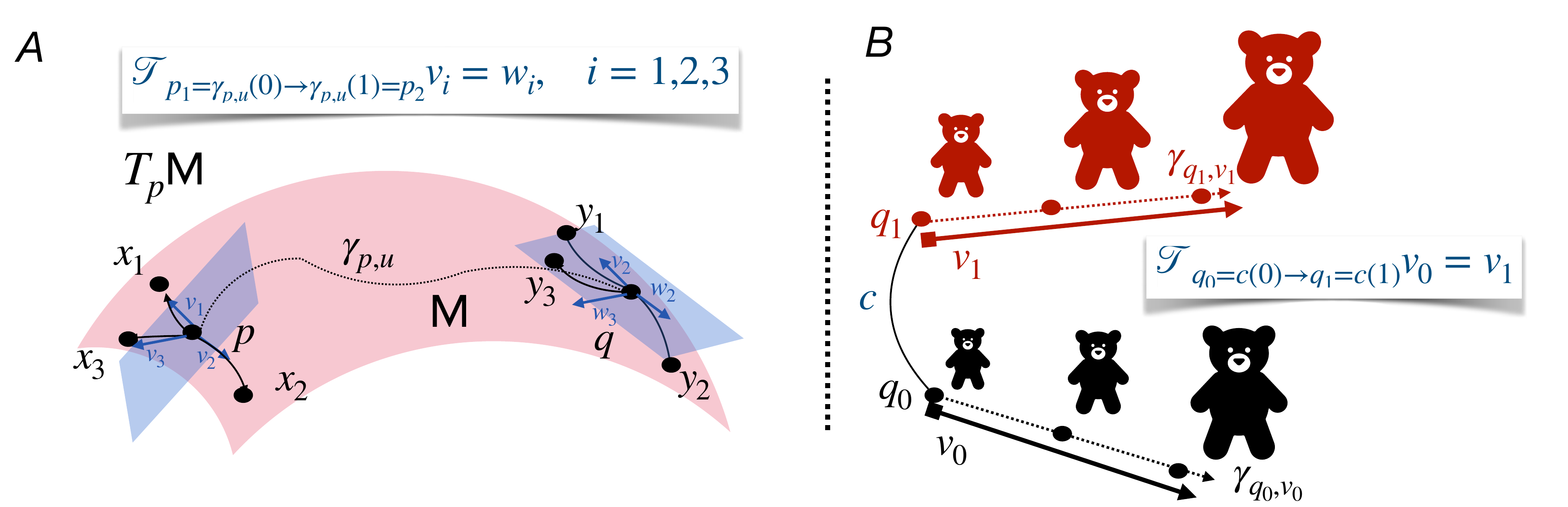}
	\centering
	\caption{
		On the left (A), the domain adaptation is performed by transporting the normal coordinates from $p_T$ to $p_S$.
		On the right (B), the growth of the black bear is transferred to the red bear using parallel transport in a manifold encoding bear shapes and colors.
		}
	\label{fig:transfer_learning}
\end{figure}

Furthermore, domain adaptation can also benefit from Riemannian metric learning using parallel transport.
For instance, in cardiac motion modeling, when a motion is estimated as a geodesic passing through $q_0$ in the source domain, we can transport this trajectory to a point $q_1$ in the target domain by transporting its velocity in parallel along the geodesic going from $q_0$ to $q_1$ \cite{guigui2022parallel}. 
This is illustrated on the right of \Cref{fig:transfer_learning} with bear shapes instead of cardiac motions for the sake of simplicity.

Tam et al.~\cite{tam2023transfer} apply a similar reasoning when working on a Riemannian space of probability distribution. 
Points in the source and target domains can be represented using normal coordinates at their Karcher means $p_S$ and $p_T$ \eqref{eq:karcher_mean}, respectively.
Then the normal representation of the target features can be transported to the source domain using the parallel transport following a trajectory going from $p_T$ to $p_S$. 
This is illustrated on the left of \Cref{fig:transfer_learning}.
In practice, to reduce the computational burden, Tam et al.~\cite{tam2023transfer} learn a statistical model (linear regression, PCA) on the normal representations in the source domain, then they transport the statistical model to the target domain.
Crucially, this transport is independant of the number of data points \cite{tam2023transfer}.

More generally, when one learns a regressor used for a downstream task, it is important to adapt the loss of your regression according to the task you target.
This idea was developed by Bansal et al.~\cite{bansal2024taskmet} using Riemannian metric learning and successfully tested on portfolio management applications.

\paragraph{Causal Inference}
Causal inference is a growing subject in the machine learning community \cite{kaddour2022causal,scholkopf2022causality,brand2023recent}.
Recently, a link was drawn between causal inference \cite{pearl1998graphs} and Riemannian geometry \cite{dominguez2023data}.
In causal inference \cite{pearl2009causal}, the outcome variable $Y$ is represented as the product of covariates $X$ and treatment $T$, such that $Y=f(X,T,U)$ where $U$ is a random exogenous factor. T
he goal is then estimate to the causal effect of the treatment.
For instance, $Y$ could indicate whether the patient has a headache, $X$ the age and the diet of the patient, and $T$ could indicate whether the patient has taken some pain medication.
Here, we might aim to measure the effect of the pain medication on the headache.
Most of the time, the treatment effect is estimated with matching methods: we hire a group of patients which will take the treatment, then we hire a control group of patients which will not take the treatment.
However, we select the second group such that each patient in the first group is paired with a patient in the control group having similar covariates.
By doing so, the covariate effect is approximately removed and we can directly observe the treatment effect by computing the difference of average between the two groups.

The causal mechanism gives structural equations (\eg~$Y=f(X,T,U)$) which can induce a geometric structure on the data \cite{dominguez2023data}.
Thus, when manipulating causal inference data, we should respect this structure.
This idea was developed by Pradier and González~\cite{pradier2024beyond}, who learn a Riemannian metric to compute the Riemannian distance between patient covariates $X$ before applying matching methods: the patients are paired if the distance between their covariates is small enough.
This method revealed to be competitive on the synthetic version of the Infant Health Development Program Dataset introduced by \cite{louizos2017causal} and the Lalonde dataset \cite{lalonde1986evaluating}.
Similar ideas have been applied to generate counterfactual explanations \cite{pegios2024counterfactual} by sampling in a Riemannian latent space.

When the treatment is given to patients in a community where their health conditions influence each other, the measurement of the treatment effect is complicated further.
Recently, Farzam et al.~\cite{farzam2024geometry} have drawn a link between the curvature of the graph related to the community and the treatment effect estimation.
They even suggest to use the Ricci curvature related to the underlying manifold to adapt the metric of the graph before performing the estimation.

In summary, since the geometric point of view on structural equations is relevant \cite{dominguez2023data,farzam2024geometry}, causal inference benefits from Riemannian metric learning by using the Riemannian distance \cite{pradier2024beyond} or more complicated objects such as the Ricci curvature \cite{farzam2024geometry}.

\section{The Mathematical Problem: How to Learn a Metric?}
\label{sec:how}
Now that we have introduced a range of applications for metric learning, we will turn back to its technical aspects, \ie~how to make the abstract problem from \Cref{sec:abstract_problem} more concrete and how to solve it.
While the choice of the objective function $\loss$ is at the crux of the problem (\Cref{sec:objectives}), the choice of the parametrization (\Cref{sec:parametric}) will deeply influence its flexibility and its computational tractability.
Interestingly, the optimization methods related to Riemannian metric learning (\Cref{sec:optimization}) are less specific to the problem than one would expect, the discussion on this issue is continued in \Cref{sec:future}.

    \subsection{Parametrization of the Metric}
    \label{sec:parametric}

The first aspect we discuss is how to parametrize the learned metric, or, in other words, which ansatz space \(\mathcal{G}_\theta \subset \mathcal{G}_\manifold\) to select for the learning problem \eqref{eq:abstract_problem}.  
As highlighted in \Cref{sec:abstract_problem}, the manifold of metrics \(\mathcal{G}_\manifold\) on a manifold \(\manifold\) is infinite-dimensional. 
Thus, to perform optimization, we must restrict our focus to a finite-dimensional subspace.  
Over the years, various approaches have been proposed for this purpose, and we will summarize and critically discuss the most significant ones.  

To achieve this, we broadly categorize the approaches into two categories: explicit and implicit.  
In \emph{explicit} approaches, a map \(\manifold \to \R^{d \times d}\) is directly learned. 
Conversely, \emph{implicit} approaches involve learning an alternative structure---such as a distance function, an embedding and so on---on the manifold, which still induces a Riemannian metric.

At the end of this section, we discuss the computational challenges related to the estimation of the Riemannian objects $\mathcal{F}(\metric)=(\dist_\metric,\Exp_\metric,\parTp_\metric,\vol_\metric)$ related to the metric.

\subsubsection{Explicit Methods}
Choosing an explicit expression for the metric can be viewed as parametrizing a function that maps to the space of symmetric positive-definite (SPD) matrices \(\mathcal{S}_{++}^n\).
This task parallels regression methods, where the function can be chosen as constant, piecewise constant, or built using kernels or neural networks.

\paragraph{Constant Metric}
The first---and most straightforward---explicit approach is also one of the most prevalent in the metric learning literature: selecting a \emph{constant} metric across the entire manifold.  
In the case of an Euclidean space \(\manifold = \R^d\), this reduces to learning a single SPD matrix \(G\). 
This approach has been widely used in distance metric learning \cite{xing2002distance,davis2007information,globerson2005metric,goldberger2004neighbourhood,guillaumin2010multiple,parameswaran2010large,shalev2004online,wang2014kernel,weinberger2009distance,xie2013multi,zhai2013heterogeneous,zhu2013point,chen2018adversarial,hsieh2017collaborative}.
It is typically framed as learning a \emph{Mahalanobis distance}:
\begin{equation}
	\dist(x,y) = \sqrt{(x-y)^T G (x-y)} = \lVert A(x-y) \rVert,
\end{equation}
where \(G = A^T A\).  

The idea of a constant metric has also been extended to other types of manifolds.
A common extension is to matrix manifolds, where the Frobenius inner product can be weighted, for example, via expressions like \(\tr(V^T G W)\) or \(\tr(G V^T W)\), or by adapting the distance functions introduced in \Cref{sec:typical_manifold}.
Such approaches have been applied to SPD matrix manifolds \cite{huang2015log,congedo2017riemannian,vemulapalli2015riemannian,zhu2018towards} and Grassmannian manifolds \cite{hamm2008grassmann,harandi2011graph,zhu2018towards}.

Another common extension targets homogeneous spaces, such as \(SO(n)\), where the metric at any point on the manifold can be obtained by deforming a reference metric defined at the identity element.
In this case, one only needs to learn a suitable metric at the reference point with the usual diffeomorphic action propagating it to the entire manifold.
This idea has been successfully applied to products of \(SO(3)\) by Kushner et al.~\cite{kushnerlearning} and to finite distributions \cite{xu2018multi,cuturi2014ground,jawanpuria2024riemannian,kerdoncuff2021metric}.

While this parametrization is computationally efficient, it has a significant limitation: it fails to capture the inhomogeneous geometry of more complex manifolds and data.

\paragraph{Piecewise Constant Metric}
This prompted researchers to explore \emph{spatially varying metrics}.  
A natural first step in this direction is to consider \emph{piecewise constant metrics}, which correspond to a non-conforming ansatz space, as the resulting metric is not globally smooth.

The most common approach for constructing piecewise constant metrics involves defining a metric at selected key points \(c_i \in \manifold\) and extending it to the entire manifold using nearest-neighbor interpolation.
This effectively assigns a constant metric within each Voronoi cell. 
Such an approach has been employed in studies such as \cite{qiu2024estimating,hauberg2012geometric}.

An alternative method involves discretizing the manifold with a (simplicial) grid and assigning a constant metric to each grid cell.
This strategy, inspired by finite element methods, was explored by Cui et al.~\cite{cui2024optimal} and Qiu et al.~\cite{qiu2024estimating}, who applied it from a computational perspective.
However, discretizing high dimensional manifolds using a grid quickly yields computational challenges.

While these methods are conceptually simple and offer flexibility, they do not yield a smooth Riemannian metric due to discontinuities at the boundaries of Voronoi cells.
The lack of smoothness at these singularities may limit their theoretical rigor and practical applicability.

\paragraph{Kernel Metric}
A conforming approach, meaning one that yields a \emph{smooth} metric, can be achiev-ed through \emph{smooth interpolation}.  
This is often accomplished using kernel-based or radial basis function interpolation.
In this framework, one selects or optimizes key points \(c_i \in \manifold\) along with corresponding metrics \(\metric_i\) defined on the tangent spaces \(T_{c_i} \manifold\).
The metric at any point \(x \in \manifold\) is then obtained as the weighted sum
\begin{equation}
	\metric_x = \sum_i \metric_i \, k(x, c_i),
\end{equation}

where \(k(x, c_i)\) is a kernel function that defines the interpolation weights.
This method has been studied extensively in the literature \cite{hauberg2012geometric,louis2019computational,sauty2022longitudinal,sauty2022riemannian,shen2019region,chen2019distance}, with Gaussian kernels being a popular choice, typically of the form \(k(x, y) = e^{-\frac{\lVert x - y \rVert^2}{\sigma^2}}\).
As with constant metrics, this interpolation approach can be extended to matrix manifolds and homogeneous spaces by leveraging their specific geometric structures.

While this method provides a flexible and moderately interpretable way to define Riemannian metrics, it may face scalability issues in high-dimensional spaces due to the computational complexity of kernel evaluations and the potential curse of dimensionality.

\paragraph{Neural Metric}
Last but not least, an approach to explicitly parametrize the metric, which is gaining increasing popularity, involves the use of \emph{neural networks}.  
Thus far, these methods have primarily been applied in Euclidean spaces, where a key challenge is ensuring that the output of the network defines an SPD matrix.

For example, Arvanitidis et al.~\cite{arvanitidis2021geometrically} address this challenge by restricting the metric to a scaled identity matrix of the form \(\lambda(x) \Idd\), where the scaling factor \(\lambda(x)\) is also tied to the data density.  
In contrast, Scarvelis and Solomon~\cite{scarvelis2023riemannian} and Bansal et al.~\cite{bansal2024taskmet} propose a more general parametrization of the metric as \(G_\theta = A_\theta^T A_\theta^{\vphantom{T}} + \eta \Idd\), where \(A_\theta \colon \R^d \to \R^{d \times d}\) is a neural network and \(\eta\) is a small positive constant to ensure positive definiteness.

Another neural network-based approach takes inspiration from the Hamiltonian formulation of Riemannian geometry (see \Cref{remark:hamiltonianflow}) by directly parametrizing the Hamiltonian, as explored in \cite{toth2019hamiltonian,chen2022learning}.
This perspective allows for learning the underlying geometric structure through dynamics.

Finally, Chadebec et al.~\cite{chadebec2020geometry, chadebec2022data}, Scarvelis and Solomon~\cite{scarvelis2023riemannian}, and van Veldhuizen~\cite{vanveldhuizen2023geometry} combine neural networks with interpolation techniques and variational autoencoder latent distribution.
In their approach, the metric is first parametrized at key points using a neural network of the form \(G_i = A_i^{\vphantom{T}} A_i^T + \eta \Idd\), and then extended across the manifold via Gaussian kernel interpolation.

While neural network-based methods benefit from the wealth of ongoing research and tools in deep learning, they remain computationally expensive and often lack interpretability compared to more traditional approaches.

\paragraph{Hand-crafted Metric}
Since learning a Riemannian metric completely can be computationally expensive and given that some problems can be solved without precise metric estimation, several studies propose to define the Riemannian metric manually based on the desired geometric properties.  

In many cases, the primary objective is to ensure that geodesics remain close to an observational dataset \(\mathcal{D} = (x_i)_i\), as real-world observations often lie on an unknown manifold \(\widetilde{\mathcal{M}} \subset \R^d\).  
To achieve this, Arvanitidis et al.~\cite{arvanitidis2016locally,arvanitidis2018latent} and Kapusniak et al.~\cite{kapusniak2024metric} suggest defining a metric \(\metric\) on \(\R^d\) as
\begin{equation}
	\label{eq:hand_crafted_metric}
	\metric_x = \left(\diag(h(x)) + \epsilon \Idd\right)^{-1}, \quad h(x) = \left(\sum_{i=1}^N (x_i^j - x^j)^2 \exp\left(-\frac{\lVert x - x_i \rVert^2}{2\sigma^2}\right)\right)_{j \in [d]},
\end{equation}
for any point \(x \in \R^d\) with coordinates \(x = (x^j)_{j \in [d]}\) where \(\epsilon > 0\) and \(\sigma > 0\) are hyperparameters to be tuned. 
With this choice, when $x$ is far from the dataset $\mathcal{D}$, $h(x)$ will be small because of the exponential term contrary to the metric's minimal eigenvalue since $\metric_x \approx \epsilon^{-1} \Idd$. 
Thus, when $x$ is far from the dataset $\mathcal{D}$, it induces large energies of displacement as the sketched in \Cref{fig:preli_scheme_opti_scheme}.
To increase the contrast, the hyperparameters are chosen such that $h(x)\approx 1 $ when $x$ is in the dataset. 
This method has been shown to be effective in tasks such as flow matching, where geodesics play a critical role \cite{kapusniak2024metric}.

Hand-crafted metrics are also relevant when the related manifold is infinite dimensional such as the space of probability measures \cite{tam2023transfer} or the group of diffeomorphism \cite{varano2017tps,germain2024Shape} to avoid computational issues.
In the same way, even in finite dimension, when data is scarce and belongs to a constrained manifold, hand-crafted metrics are still relevant \cite{vanveldhuizen2023geometry,jeong2024deep}. 
Designing a metric by hand is also performed in implicit methods, but they distinguish themselves from explicit form by learning other objects used in the process.

\subsubsection{Implicit Methods}
In implicit methods, the metric is not learned as a matrix field \(\manifold \to \R^{d \times d}\) but derived from different objects.
Compared to explicit metric parametrizations, implicit methods can be more computationally efficient while still offering significant flexibility.

\paragraph{Pullback Metric}
One of the most popular implicit approaches is the use of a \emph{pullback metric}.
In this framework, a (smooth) map \(f \colon \manifold \to \mathcal{N}\) is learned from the original manifold \(\manifold\) to another Riemannian manifold \(\mathcal{N}\) and the metric on \(\manifold\) is induced by the pullback, \ie

\begin{equation}
	\label{eq:pullback_metric}
	\metric_x(v, w) = \metric_{f(x)}\left(\dd_x f \cdot v, \dd_x f \cdot w\right),
\end{equation}
where \(\dd_x f\) denotes the differential of \(f\) at point \(x\).
This pullback metric is valid as long as \(f\) is locally injective.
The map \(f\) can take various forms depending on the application.
In the simplest case, \(f\) is a linear map, which brings us back to the scenario of Mahalanobis distance learning.
More commonly, nonlinear forms of \(f\) are used including kernel methods \cite{huang2014learning,mcfee2011learning,tosi2014metrics,gruffaz2021learning} and neural networks \cite{frenzel2019latent,kalatzis2020variational,louis2019riemannian,zhou2021domain,braunsmann2024convergent}.

A frequent application of pullback metrics involves autoencoder architectures \cite{yessenbayev2024use,warburg2024bayesian,syrota2024decoder,sun2025geometry,pegios2024counterfactual,tosi2014metrics,kalatzis2020variational,beikmohammadi2021learning,acosta2023quantifying,braunsmann2024convergent} where the metric can be induced by either the encoder or the decoder depending on the task requirements.
When $f\sim P$ is random or not unique in the context, relying on the expected value $\mathbb{E}_{f\sim P}(\dd_x f)$ instead of $\dd_x f$ is an interesting extension account for the probabilistic nature of the space geometry \cite{tosi2014metrics,fasina2023neural,syrota2024decoder,rozo2025riemann}.
Arvanitidis et al.~\cite{arvanitidis2022prior} designed the metric according to a learned prior in the latent space and not using the pullback to gain robustness while still having a similar structure.

While some method directly explicitly define the pullback metric \cite{gruffaz2021learning,tosi2014metrics,louis2019riemannian,kalatzis2020variational,braunsmann2024convergent}, others, such as \cite{bronstein2010data,huguet2022manifold,huang2015projection,harandi2017dimensionality,mcfee2011learning,kedem2012non,baktashmotlagh2013unsupervised,huang2014learning,fathi2017semantic,duan2018deep,cai2017semantic,chen2019distance,ein2018learning,he2018triplet} focus on learning a local approximation of the distance using the map-induced distance
\begin{equation}
	\dist_f(x, y) = \lVert f(x) - f(y) \rVert.
\end{equation}
Note that \(\dist_f\) is only a local approximation of the geodesic distance, as it does not account for the full geometry of the manifold as explained in \Cref{remark:local_approximation}.

In some cases, the pullback metric is replaced with a \emph{pushforward metric}, where a diffeomorphism \(\tilde{f} \colon \mathcal{N} \to \manifold\) is learned instead. 
This approach induces a pushforward structure in both the input and output spaces \cite{gruffaz2021learning,dekruiff2024pullback,diepeveen2024pulling,diepeveen2024score}.
An advantage of pushforward metrics is that key Riemannian objects such as the exponential map, the logarithm map, and parallel transport often have closed-form expressions. 
These formulas are detailed in \cite{diepeveen2024pulling}.

This approach is computationaly feasible, quite flexible and interpretable as long as the function $f$ shares these property.

\paragraph*{Curvilinear Coordinate System}
Instead of directly modeling the smooth map \(f\), Chen et al.~\cite{chen2019curvilinear} and Zhou et al.~\cite{zhou2021domain} propose learning a \emph{curvilinear coordinate system}.  
This approach corresponds to identifying the principal geodesics of the target metric.
The distance between points is then computed by projecting them onto these principal geodesics---\ie~the curvilinear coordinate system---yielding a local approximation of the Riemannian distance.

This original idea has shown success in distance metric learning \cite{chen2019curvilinear}.
However, its application to general Riemannian metric learning remains limited, as this would require the computation of more complex geometric objects, such as parallel transport, to fully exploit the underlying manifold structure.

\paragraph*{Graph-based Approach}
The final implicit method for encoding a metric that we consider is the \emph{graph-based approach}.
In this approach, it is assumed that sparse measurements or approximations of distances on the manifold are encoded through the edges of a graph.
Given this graph, the distance between points can be approximated by computing the shortest paths on the graph, which in turn induces a metric \cite{benmansour2010derivatives,farzam2024geometry,heitz2021ground,li2023geometry,peltonen2004improved}.

Such graph-based methods are often used to learn distances in a non-parametric manner, based on the assumption that the dataset is isometrically embedded in an ambient space.
The nearest-neighbor graph is then computed leading to algorithms like Isomap \cite{tenenbaum2000global} and its extensions such as \cite{budninskiy2019parallel}.

\subsubsection{Computing Riemannian Objects}
\label{sec:computing}
Once one has parametrized the Riemannian metric $\metric$, one has to compute the Riemannian objects of interest $\mathcal{F}(\metric)=(\dist_\metric,\Exp_\metric,\parTp_\metric,\vol_\metric)$ as explained in \Cref{sec:abstract}.
Here, we provide a few elaborations on how to do this.

\paragraph*{Constant Metric}
For the case, that $\metric \equiv G$, where $G$ is a symmetric positive definite matrix and we are working on Euclidean space, the associated distance $\dist_\metric$ corresponds to the Mahalanobis distance. 
In this case, geodesics are straight lines, the parallel transport reduces to the identity map, and the metric volume remains constant at $\sqrt{|\det(A)|}$.
This again highlights the lack of nonlinear structure provided by a constant metric.

\paragraph*{Non-constant Metric}
For nonlinear explicit forms, no universally accepted gold standard currently exists.  
The metric volume $\vol_\metric$ requires computing a determinant at each point $x \in \manifold$.  
The exponential map $\Exp_\metric$ and the parallel transport $\parTp_\metric$ can be approximated by numerically integrating the corresponding second-order differential equations, see \Cref{sec:preliminaries}.

However, in high-dimensional settings, alternative strategies are often preferable for estimating parallel transport. 
In infinite-dimensional spaces, Schild's ladder \cite{ehlers1972geometry,lorenzi2011schild} and the Pole ladder \cite{marco2013parallel} provide approximations of parallel transport, particularly when evaluating the exponential map and logarithm is computationally efficient, as is the case for Lie groups.  
When computing the logarithm is not tractable, the fanning scheme offers an alternative approach approximating parallel transport using only the exponential map \cite{younes2007jacobi,louis2017parallel}. 
Its computational complexity scales as $d^2$ where $d$ is the dimension.

The computation of the distance $\dist_\metric(x,y)$ involves solving the minimization problem \eqref{eq:dist_as_geodesics}, which is equivalent to estimating a shortest geodesic given the endpoints $x,y \in \manifold$.  
This task is particularly challenging, and three general approaches can be employed for approximation:

\begin{enumerate}
  \item \textbf{Shooting method.} The exponential map is computed by numerically integrating the geo-desic equation and the distance is estimated as  
  \begin{equation}
    \dist_\metric(x,y) = \sqrt{\metric_x(\log_x(y), \log_x(y))}
  \end{equation}
  by approximating the logarithm map:  
  \begin{equation}
    \log_x(y) = \argmin_{v \in T_x\manifold} L(\Exp_x(v), y),
  \end{equation}
  where $L$ is a loss function often called data-attachment term measuring the discrepancy between the points.  
  Notably, this method also yields the logarithm $\log_x(y)$, which can be useful for representation tasks such as computing a Karcher mean.  
  Although relatively straightforward to implement, its computational cost can be significant due to the integration of the geodesic equation and the need to differentiate its solution.
  
  \item \textbf{Geodesic discretization.} Another approach involves computing (shortest) geodesics by discretizing in time the curve $\gamma$ and its associated path energy, as defined in \eqref{eq:path_energy}.
  This leads to
  \begin{align}
    \left(\hat{\gamma}\left(\frac{k}{N}\right)\right)_{k \in [N]} = \argmin_{(x_k)_{k \in [N]}} \sum_{k=0}^{N} \tilde\dist_\metric(x_k, x_{k+1})^2\eqsp,
  \end{align}
  where \(\tilde\dist_\metric(x_k, x_{k+1})^2\) is a local approximation of the squared distance, \eg \newline \(\tilde\dist_\metric(x_k, x_{k+1})^2 \approx \metric_{x_k}(x_{k+1} - x_k, x_{k+1} - x_k)\).
  Here, $N$ must be sufficiently large to ensure a realistic solution \cite{heeren2014exploring,arvanitidis2019fast} and the convergence of this approach for \(N\) going to infinity was investigated by Rumpf and Wirth~\cite{rumpf2015variational}.
  The minimization can be performed in various ways. 
  A recent fast and robust approach suggests using Gaussian processes \cite{arvanitidis2019fast} combined with fixed-point iterations to avoid the computational overhead of Jacobian processing.  
  While such methods may be less precise than the shooting method, in some cases shooting methods are not feasible especially in high dimensions.
  
  \item \textbf{Neural network regression.} When only a rough approximation of the distance or geodesic is required, a neural network $\phi_\eta \colon \manifold \times \manifold \times [0,1] \to \manifold$ parameterized by $\eta$ can be trained to produce geodesics between given points.  
  Concretely, in \cite{kapusniak2024metric,sun2025geometry}, geodesics are parameterized as  
  \begin{equation}
      \gamma_\eta(x,y,t) = (1-t)x + ty + t(1-t) \phi_\eta(x,y,t),
  \end{equation}
  where $\eta$ is estimated by minimizing a stochastic versions of the path energy \eqref{eq:path_energy}:
  \begin{equation}
      \eta \mapsto \mathbb{E}_{x,y \sim P_{\mathcal{D}} \otimes P_{\mathcal{D}}, t \sim \mathcal{U}([0,1])} \left( \metric_{\gamma_\eta(x,y,t)} (\dot{\gamma}_\eta(x,y,t), \dot{\gamma}_\eta(x,y,t)) \right),
  \end{equation}
  using stochastic gradient descent.
  Here, $P_{\mathcal{D}}$ represents the empirical distribution associated with noisy observations $\mathcal{D} = (x_i)_i$ on the manifold.  
  
  In practical applications, the metric is typically hand-crafted as in \eqref{eq:hand_crafted_metric} to ensure that geodesics remain close to the manifold approximated by the dataset.  
  This method has proven to be both practical and sufficiently accurate to outperform state-of-the-art techniques in trajectory inference \cite{kapusniak2024metric,sun2025geometry}.  
  Additionally, it has been extended using splines in the context of optimal transport \cite{pooladian2024neural}.  
  
\end{enumerate}

\paragraph*{Pullback Metric With Diffeomorphisms}
As mentioned earlier, if the pullback metric \eqref{eq:pullback_metric} is defined via a diffeomorphism $f$ on the ambient space \cite{gruffaz2021learning,dekruiff2024pullback,diepeveen2024pulling,diepeveen2024score}, then all objects in $\mathcal{F}(\metric)$ admit a closed-form expression as presented in \cite{diepeveen2024pulling}.  
This modeling approach is particularly useful when working in the observational space. However, estimating $f$ becomes challenging in high-dimensional settings.  
For this reason, variational autoencoders (VAEs) are often preferred in high-dimensional spaces, although they are not diffeomorphic mappings, since the latent space has a lower dimension than the observational space.

\paragraph*{Graph-based Metrics}
When using a graph-based structure, geodesics are estimated via shortest path algorithms such as Dijkstra's algorithm \cite{mirebeau2017automatic,benmansour2010derivatives,peltonen2004improved} or approximations based on the heat equation \cite{heitz2021ground}. 
The same holds for distance computations.  
Parallel transport can be computed using ad hoc methods based on local Principal Component Analysis (PCA) \cite{budninskiy2019parallel}.  
Curvature is approximated using graph-equivalent definitions \cite{farzam2024geometry,li2023geometry}.

    \subsection{Objectives}
    \label{sec:objectives}
    
After having discussed the parametrization of the learned metric, we now turn to another central aspect of making the abstract problem \eqref{eq:abstract_problem} concrete: the choice of objective.
It is, of course, the central ingredient to define \emph{which} Riemannian metric we want to obtain in the end restricted by the space of possible metrics defined above and the minima achievable by the chosen optimization algorithm discussed in the next section.

The objective is usually chosen with some application of the metric in mind. 
We will thus first classify the objective broadly into two main categories: objectives based on regression and objectives based on classification.
This follows, on the one hand, the very common classification of machine learning problems and, on the other hand, also mirrors the historical development of metric learning outlined in the introduction.
There are some objectives that do not fit perfectly into these two categories and we will discuss them in a third subsection afterwards.

\subsubsection{Classification}
\label{sec:objectives_classification}
As explained before, the original motivation leading to the introduction of distance metric learning was to improve the performance of \(k\)-nearest neighbor classification. 
As the term \emph{nearest} suggests, the distance between samples is a central ingredient and, thus, a wide range of objectives have been suggested that are based on this application.
They can be carried over to Riemannian metric learning via the geodesic distance derived from the metric, which is the way will formulate them here.
We will focus on a subset of important examples which will not be exhaustive as this would go beyond the scope of our review.
For a more detailed overview on these losses, we refer to existing reviews focused on distance metric learning, such as the one by Kaya and Bilge~\cite{kaya2019deep}. 

\paragraph{Contrastive}
The first set of objectives based on classification we consider are usually referred to as \emph{contrastive losses} \cite{chen2018adversarial,chen2019curvilinear,davis2007information,fathi2017semantic,globerson2005metric,guillaumin2010multiple,hamm2008grassmann,harandi2011graph,harandi2017dimensionality,hauberg2012geometric,huang2014learning,huang2015log,huang2015projection,kedem2012non,wang2018multi,warburg2024bayesian,xing2018pan,duan2018deep,bauer2024elastic}.
Here, one assume two sets as input: a set \(S_p \subset \mcx \times \mcx\) of examples that should be classified as similar (\eg~have the same label) and another set \(S_n \subset \mcx \times \mcx\) of examples that should be classified as different.
These sets, especially the latter, are typically assumed to contain ``challenging'' examples for a classification algorithm, \ie~such examples that are close to an envisioned decision boundary (in the Euclidean norm).
Hence, the objective in metric learning is to achieve a clearer separation of these examples, \ie~to \emph{maximize} the distance for elements of \(S_n\), while keeping similar points close, \ie~to \emph{minimize} the distance for elements of \(S_p\).
Various different objectives have been proposed that formalize this idea, for example, as
\begin{equation}
	\loss(\metric) = \sum_{(p,p^\prime) \in S_p} \dist_\metric(p,p^\prime) - \sum_{(q,q^\prime) \in S_n} \dist_\metric(q,q^\prime)
\end{equation}
in \cite{huang2014learning} or as 
\begin{equation}
	\loss(\metric) = \sum_{(p,p^\prime) \in S_p} \dist_\metric(p,p^\prime) + \sum_{(q,q^\prime) \in S_n} \dist_{\metric^{-1}}(q,q^\prime),
\end{equation}
in \cite{zhu2018towards} where the inverse of the (local) metric is considered for negative pairs.
In a slightly different direction, also placing constraints on the distances for one set, \eg~demanding that the distance for all examples in \(S_p\) is below a certain threshold, while optimizing the distance for the other has been considered \cite{vemulapalli2015riemannian}.
This amounts to replacing one of the sums in the above objectives with an indicator function which is zero if a pair fulfills the imposed threshold and is infinity otherwise.
Various other variation of this idea have been proposed and we refer, for example, to \cite{kaya2019deep} for a more complete overview.

\paragraph{Triplet}
Another set of objectives, that is typically referred to as \emph{triplet losses} \cite{annarumma2018deep,cai2017semantic,chen2019distance,ein2018learning,he2018triplet,hermans2017defense,hsieh2017collaborative,hu2015deep,liu2018efficient,wang2019centroid,zhang2021curvilinear,duan2018deep}, assumes that similar information is given in the form triplets \(S_t \subset \mcx \times \mcx \times \mcx\).
Here, each triplet \((q, q_p, q_n)\) contains an anchor point \(q\), a positive example \(q_p\) to be classified as similar, and a negative example \(q_n\) to be classified as different.
Based on these triplets, most commonly 
\begin{equation}
	\loss(\metric) = \sum_{(q, q_p, q_n) \in S_t} \dist_\metric(q,q_p) + \max(0, 1 + \dist_\metric(q,q_p) - \dist_\metric(q,q_n))
\end{equation} 
is used as objective, \eg~in \cite{weinberger2009distance}. 
It encourages the positive example \(q_p\) to be close while pushing the negative example (also called impostor) \(q_n\) \emph{further} away than \(q_p\) by a margin (which is one here).
The idea behind this objective is to ensure a ``clean'' neighborhood for the anchors, \ie~such that every point with a different label is farther away than points with the same label as the anchor. 
Thus a (k-)nearest neighbor classifier would correctly classify the anchor based on its neighborhood.
Again, over time, variations of this objective have been proposed and we refer, for example, to \cite{kaya2019deep} for a more complete overview.

\paragraph{Others}
Since classification-based objectives have been the main types of objective considered for distance metric learning, naturally many more have been developed over the course of time and we want to mention a few ideas motivating these other objectives. 
One interesting direction is to achieve unsupervised or at least semi-supervised objectives instead of the supervised approaches discussed above.
Another direction is to model more non-local interaction between points, \eg~considering not just triplets but larger tuples or entire cluster of points, primarily to speed-up the convergence of descent algorithms, which otherwise might need a lot of iterations due to only local changes affected in each step \cite{mcfee2011learning,goldberger2004neighbourhood,wang2017deep}.
A common challenge that most of these approaches share with the one we have explained is the need to sample points from the dataset in a specific way \cite{kaya2019deep,qiu2024estimating}.
For example, for the triplet loss, we need the anchors to be sufficiently densely sampled, and for each anchor, sufficiently many negative examples available to control their neighborhood \cite{weinberger2009distance}.

\subsubsection{Regression}
In this second category of objectives, we assume to be given a more geometric type of supervision to learn the desired metric, \ie~observations of some continuous object derived from the metric instead of a classification.
This could be scalar quantities, as \eg~the distance between points, geometric quantities such as trajectories on the manifold, or even just the dataset itself describing its distribution.
Below, we will discuss examples for these different types of supervisions.
This second category of objectives is usually more closely connected to Riemannian metric learning in the literature, although, as argued before, this is not necessary conceptually.

\paragraph{Distribution}
The most fundamental form of supervision is given just by (potentially noisy) samples \(\mathcal{O} \subset \manifold\) on the manifold, which can be thought of as observations of its density.
Thus, this first set of objectives broadly aims to learn a metric such that the corresponding volume form matches this density, which can be turned into concrete objectives in various ways.

To achieve this goal, Arvanitidis et al.~\cite{arvanitidis2016locally} construct a diagonal metric where distances are smaller in areas with more samples and larger far away from them using the inverse of the local diagonal covariance matrices.
They show empirically that a corresponding Riemannian normal distribution then reasonably approximates the data density.
In \cite{arvanitidis2018latent}, they extend this idea to an expected pull-back metric on the latent space of variational autoencoders, where the variance of the decoder is optimized to be large away from the given samples. 
Finally, in \cite{arvanitidis2021geometrically}, they optimize a metric to give constant distances in dense areas while growing when moving away from them and combine this with the variance estimation of \cite{arvanitidis2018latent} to pull the metric back to the latent space of a variational autoencoder.
This first part of this approach is also used by Kapusniak et al.~\cite{kapusniak2024metric}.

Another approach is followed by Chadebec et al.~\cite{chadebec2020geometry,chadebec2022data}, who train a variational autoencoder using the common ELBO loss to approximate the data density. 
However, the distribution on the latent space is assumed to be given via a learned Riemannian metric and the sampling to evaluate the objective is performed using Riemannian Hamiltonian Monte Carlo.
A similar idea is followed by Kalatzis et al.~\cite{kalatzis2020variational}, who use a Riemannian Brownian motion as prior for the distribution in the latent space using a pulled-back metric.
Toth et al.~\cite{toth2019hamiltonian} and Sauty and Durrleman~\cite{sauty2022riemannian} combine the idea of approximating the data distribution using a variational autoencoder with approximating given trajectories as will be explained below.

\paragraph{Distance}
Starting from distance observations, we assume having insights into the global geometric structure of our manifold and aim to derive a Riemannian structure matching it.
Concretely, we assume to have a dataset \(\mathcal{O} \subset \manifold \times \manifold \times \R_{\geq 0}\) consisting of pairs of points \((p,q)\) on the manifold along with a (possibly noisy) observation \(d\) of the distance between them.
Then the objective usually turns into a basic least squares estimation, \ie
\begin{equation}
	\loss(\metric) = \sum_{(p,q,d) \in \mathcal{O}} \lvert d - \dist_\metric(p,q)\rvert^2.
\end{equation}
If one has a notion of density of the samples associated with them, \ie~some form of weights, these can of course also be integrated into the objective.
This approach to regress distances has, for example, been investigated in \cite{dekruiff2024pullback,diepeveen2024pulling,diepeveen2024score,frenzel2019latent,qiu2024estimating,braunsmann2024convergent}.

\paragraph{Trajectories}
For our last set of regression-based objectives, we assume that we are not just given observations of distance but of entire trajectories on the manifold which we want to model as geodesics under our learned metric.
To this end, we assume that we are given $K$ trajectories \((x_k^{j})_{k\in [K],j\in[T_k]}\) at variable time points \((t_k^j)_{k=1 \in[K],j\in[T_k]}\) where $T_k$ is the number of observations in trajectory $k$.
Based on this data, we can introduce a formal objective for the regression problem via the set of all geodesics \(\mathcal{T}_\metric\) for a given metric \(\metric\) as
\begin{equation}
	\label{eq:formal_geodesic_loss}
	\loss(\metric) = \sum_{k=1}^K \inf_{\gamma \in \mathcal{T}_\metric} \sum_{j=1}^{T_k} \lVert x_k^j - \gamma(t_k^j) \rVert^2\eqsp .
\end{equation}
Since optimizing over the set of all geodesics for a given metric is rarely achievable this formulation is mostly of educational value.
Hence, a central challenge is how to parametrize a sufficiently expressive subset of geodesic that allows to approximate a objective of the form \eqref{eq:formal_geodesic_loss} in a tractable manor.

To this end, Louis et al.~\cite{louis2019riemannian} consider an autoencoder setup where the metric on the data manifold is given by the push-forward of the Euclidean metric on the latent space (cf.~\Cref{sec:parametric}). 
This means geodesics on the data manifold are given by the images of linear trajectories in the latent space, which are much more tractable to work with.
Furthermore, they assume that all observed trajectories are parallel to each other (and a coordinate axis), which means that each trajectory is described by its constant speed, its starting time relative to the observations, and its offset to the coordinate axis.
These quantities are then optimized together with the parameters of the autoencoder determining the Riemannian metric considering a objective of the form \eqref{eq:formal_geodesic_loss} and an additional regularization on them.
This idea has been extended by Sauty and Durrleman~\cite{sauty2022longitudinal} to variational autoencoders also modeling the data distribution (see above).
Here, the regression objective is obtained by comparing points in the latent space whereas Louis et al.~\cite{louis2019riemannian} compare points in the original data space after decoding.

This idea can be generalized to nonlinear \emph{\(\exp\)-parallel curves} \cite{schiratti2017bayesian} (see \Cref{sec:applications_riemannian_metric_learning}), which are not necessarily geodesics. 
Sauty and Durrleman~\cite{sauty2022riemannian} employ this idea while directly learning a metric on the data manifold. 
Thus, they approximate the geodesic and the parallel curves using a Hamiltonian shooting approach.
In contrast, Gruffaz et al.~\cite{gruffaz2021learning} obtain the metric using a diffeomorphic deformation of a reference manifold (usually the Euclidean space) with known closed forms for the geodesic and parallel curves.

Aiming for a larger candidate set, Toth et al.~\cite{toth2019hamiltonian} consider general geodesics in the latent space of a variational autoencoder parametrized by their starting point and direction which are used in a Hamiltonian shooting.
Finally, Kushner  et al.~\cite{kushnerlearning} assume that they are given coarsely sampled discrete trajectories and aim to upsample them using geodesics.
Hence, they parametrize candidate geodesics by tangent vectors at given points on the manifold.

\subsubsection{Others}
In this last subsection, we discuss some works performing Riemannian metric learning with objectives that do not fall into the aforementioned categories. 

\paragraph{Optimal Transport}
As explained in \Cref{sec:typical_manifold}, one can learn a metric on the space of probability measures on a manifold \(\manifold\) by learning a Riemannian metric on this base manifold.
Hence, a range of works as focused on objectives for Riemannian metric learning that incorporates optimal transport based on it.

Considering the classification of histograms, Cuturi and Avis~\cite{cuturi2014ground} adapt their objective from the contrastive losses in distance metric learning and arrive at the loss \(\loss(\metric) = \sum_{(\mu, \nu) \in S_p} \Wasserstein_2(\mu, \nu) - \sum_{(\mu, \nu) \in S_n} \Wasserstein_2(\mu, \nu)\), where \(\Wasserstein_2\) is the Wasserstein distance induced by \(\metric\) as defined in \eqref{eq:Wasserstein}.
In contrast, Zhou et al.~\cite{zhou2021domain} minimize the Wasserstein distance between subsets of data points (representing different source domains in a domain adaption setup) while combining it with a triplet loss (see \Cref{sec:objectives_classification}) over all data points.
Likewise, Kerdoncuff et al.~\cite{kerdoncuff2021metric} optimize the Wasserstein distance between two sets of data points, one with labels and one without, with an additional penalty on the transport plan preventing two differently labeled points being sent to the same target.
They also combine this with classification-based losses on the points with known labels.

Aiming towards the regression of measures, Scarvelis and Solomon \cite{scarvelis2023riemannian} consider given pairs \((\mu_k, \nu_k)_k\) of probability distributions that are assumed to be consecutive times steps of (discrete) trajectories and aim to minimize their Wasserstein distance, \ie~\(\loss(\metric) = \sum_{k} \Wasserstein_1(\mu_k, \nu_k),\) where \(\Wasserstein_1\) depends on \(\metric\) as before.
To prevent the norm of the metric becoming arbitrarily small, they include an additional regularization penalizing the norm of the inverse metric tensor along linear interpolations between samples of the given distributions.
Similarly, Heitz et al.~\cite{heitz2021ground} assume that the supervision for the metric learning comes in form of complete discrete trajectories of measures that should be well-approximated by optimal transport regression. 
Hence, they adapt the objective \eqref{eq:formal_geodesic_loss} to interpolations of measures with fixed given endpoints for the trajectories.
This means they have to solve optimal transport interpolation problems to evaluate their objective for which they use entropy-regularized optimal transport \cite{peyre2019computational}. 

Let us note that these objectives based on optimal transport could also be considered as Riemannian metric learning on the space of probability measure with the objectives introduced above. 
For example, the loss of Cuturi and Avis~\cite{cuturi2014ground} is simply a contrastive loss from before with the Wasserstein distance.
However, we chose to consider these objectives for the base metric based on optimal transport.
This is because, to the best of our knowledge, there is thus far no viable approach to directly learn a metric on the space of probability measures forgoing the base metric on the one hand and, on the other hand, these objectives are sometimes combined with objectives that do not involve optimal transport.

\paragraph{Optimization}
Finally, a few works focus on the use of the Riemannian metric in optimization applications and design objectives accordingly.
For example, Bansal et al.~\cite{bansal2024taskmet} learn the metric using a bilevel optimization problem, where the learned metric is used in the lower level problem to asses the prediction accuracy of a regression model while the overall objective (\ie~upper level problem) is to optimize the performance of this model on a different task. 
Such problems arise, \eg, in reinforcement learning when the overall objective measures the quality of the model's downstream value predictions while the regression problem concerns next state prediction.

In another direction, Cui et al.~\cite{cui2024optimal} minimize the Poincaré constant implied by the Riemannian metric for a given distribution on the manifold.
The resulting metric is then used in Riemannian Langevin dynamics to achieve efficient sampling from the distribution as its convergence rate is given via the Poincaré constant.

    \subsection{Optimization Methods}
    \label{sec:optimization}
Having chosen the set of metrics over which to optimize and the objective to be minimized, we have to also choose a practical algorithm to find an approximate solution.
Over time, numerous approaches have been proposed for this and we give a overview of them in this last section on the technical aspects of Riemannian metric learning.
Since the chosen optimization approach often amounts to the application of a standard algorithm, we will keep the overview brief and give some remarks about common things to note when applying the various algorithms to the metric learning problem.

\paragraph{Closed Form Formulas}
The best-case scenario from a performance point of view is that we have a known closed form of the minimizer of the learning problem given the training data, \eg~in form of an eigendecomposition.
This is, for example, the case for certain approaches using a constant metric and classification-based losses as introduced in the previous section as the problems are convex with a known global solution.
This has been used in \cite{hamm2008grassmann,harandi2011graph,hauberg2012geometric,zadeh2016geometric,zhu2018towards} among others.
For other approaches, such as \cite{arvanitidis2016locally}, this is the case as the metric is defined primarily to follow the density of the data and constructed from the beginning with a formula in mind.
The metric can also be known if it is constructed via the pull-back along an otherwise constructed map, \eg~an autoencoder, which was used in \cite{frenzel2019latent,tosi2014metrics}.
However, the number of problems that admit a known closed form solutions is small and especially it is rarely true for non-linear problems.

\paragraph{Gradient Descent Methods}
Thus, in the absence of closed form solutions, most works turn towards iterative descent algorithms to solve the optimization problem.
Here, the most fundamental approach is to apply gradient descent or its stochastic extension which was used in \cite{bansal2024taskmet,chen2019curvilinear,globerson2005metric,goldberger2004neighbourhood,kerdoncuff2021metric,zhou2021domain,shalev2004online,kedem2012non,xing2002distance,zhang2021curvilinear,parameswaran2010large,myers2022regression}.
To speed-up the convergence of the descent, some works, such as \cite{arvanitidis2021geometrically,chadebec2020geometry,chadebec2022data,kapusniak2024metric,louis2019computational,toth2019hamiltonian,sauty2022longitudinal,sauty2022riemannian,sauty2023multimodal,scarvelis2023riemannian,braunsmann2024convergent}, use acceleration techniques such as Nesterov's acceleration or the Adam algorithm.
Others \cite{heitz2021ground,shen2019region} use L-BFGS to approximate the Hessian of the objective and achieve faster convergence.

These iterative methods typically face two challenges: (i) computing the gradient, and (ii) ensuring that the learned metric is positive definite for explicit parametrizations.
The first challenge traditionally required manual computation of the gradient of the objective which has been mostly replaced by the usage of automatic differentiation frameworks in more recent approaches.
Furthermore, the objective might not admit a classical gradient, \eg~if it involves solving an optimization problem as in the case of optimal transport-based objectives.
Thus, some works explicitly use subgradient methods \cite{cuturi2014ground,kedem2012non,xie2013multi,parameswaran2010large}.
The second challenge is often solved via the parametrization of the metric (see \Cref{sec:parametric}) and thus mostly comes up in the optimization for constant metrics.
In these cases, it is typically \cite{globerson2005metric,xing2002distance} solved by projected gradient descent, \ie~alternating a gradient step with projecting the metric matrix onto cone of positive definite matrices.

\paragraph{Structured Optimization}
Instead of applying acceleration techniques, another range of works focuses more on the concrete structure of the considered optimization problem.
For example, \cite{davis2007information,huang2015log,vemulapalli2015riemannian} apply Bregman iterations to the problem consisting of a convex objective with convex constraints.
Alternatively, some problems can be reformulated as a semidefinite program which allows to use corresponding specialized optimization techniques, \eg~ones developed to train SVMs.
This was used in \cite{mcfee2011learning,wang2014kernel,weinberger2009distance,zhu2013point} for example.
Furthermore, some methods, \eg~\cite{huang2014learning,zhai2013heterogeneous}, split the parameters of the metric into different subsets and then alternate optimizing over these subset, which is usually the case when for one subset a more structured optimization can be applied. 

Another subset notably uses the Riemannian structure of the set of SPD matrices itself in the optimization by turning to Riemannian optimization methods.
These methods use the Riemannian gradient of the objective in the tangent space of the current iterate and then apply an approximation of the exponential map to perform line search along it or a derived descent direction.
This has been applied in \cite{harandi2017dimensionality,huang2015projection,yger2015supervised} to avoid degeneracies of the metric and speed-up the convergence compared to methods such as projected gradient descent.

All of these methods require sufficient insight into the structure of the problem, either in terms of the objective or in terms of the parametrization of the metric. 
This is the reason why they have been applied mostly in approaches considering constant metrics or kernel methods.
The general theory of Riemannian metric learning is thus far not developed enough to produce structured methods regardless of the chosen parametrization which can also applied, \eg, to neural network-based approaches, see also \Cref{sec:future}.

\section{Future Directions}
\label{sec:future}

This section emphasizes promising avenues for future research in Riemannian metric learning, including:
\begin{itemize}
	\item \textbf{Statistical Guarantees for Nonlinear Metric Learning.}
	Developing rigorous statistical and optimization frameworks to provide theoretical guarantees and performance bounds for Riemannian metric learning methods.
	\item \textbf{Geometry-Aware Optimization Methods.}
	Designing optimization techniques that account for the intrinsic geometric properties of the space of Riemannian metrics, thereby improving convergence and efficiency.
	\item \textbf{Enhanced Software and Methodological Tools.}
	Creating new computational tools and algorithms aimed at improving the tractability, scalability, and practical usability of Riemannian metric learning approaches.
	\item \textbf{Discovering New Applications.} As Riemannian metric learning tools become more accessible, their applications will expand. 
	In particular, leveraging Ricci curvature could enhance optimization, generalization, and sampling efficiency.
\end{itemize}

\subsection{Theoretical Research Avenues}
Theoretical researcher in statistics, optimization and geometry are challenged to contribute to Riemannian metric learning.
\paragraph{Statistical Guarantees for Nonlinear Metric Learning}

Theoretical advancements in the field of metric learning remain limited compared to the volume of empirical studies in the field.
However, there is a notable recent trend toward theoretical and statistical investigations, particularly in the context of preference systems and metric learning \cite{bellet2015robustness,verma2015sample,kleindessner2014uniqueness,mason2017learning,canal2022one,wang2024metric,chen2024pal}.  

In \cite{bellet2015robustness,verma2015sample}, the authors establish foundational results, addressing robustness \cite{bellet2015robustness} and sample complexity \cite{verma2015sample} for learning Mahalanobis distances under various loss functions.
Building on these works, Mason et al.~\cite{mason2017learning} refine the sample complexity analysis by focusing on low-dimensional metrics, particularly under regularization constraints on the learned matrix and when employing a triplet loss.  

More recently, Canal et al.~\cite{canal2022one}, Chen et al.~\cite{chen2024pal}, and Wang et al.~\cite{wang2024metric} have explored the intersection of Mahalanobis metric learning and user preference modeling.
In these studies, the objective is to jointly learn a Mahalanobis metric and user embeddings \((u_j)\) in the space, where a user's preference for \(x_1\) over \(x_2\) is encoded as \(||x_1 - u|| < ||x_2 - u||\).
Initial theoretical results were presented in \cite{canal2022one} and later refined by \cite{wang2024metric}, which introduced low-rank assumptions to improve the learning process.
Furthermore, \cite{chen2024pal} extended these ideas to pluralistic alignment in fine-tuning large language models (LLMs), further motivating theoretical advancements.  

The next significant challenge lies in addressing the non-linear case of metric learning, such as using Riemannian metrics parameterized by neural networks or non-linear pushforward maps.
Progress in this direction would benefit from the development of prior distributions and low-rank assumptions tailored to Riemannian metrics.
For instance, Arvanitidis et al.~\cite{arvanitidis2022prior} propose learning a prior in the latent space of a variational auto-encoder to derive a Riemannian metric that better accounts for latent data distribution.
However, this approach is not yet well-supported by theoretical guarantees. 

In the same way, some theory related to the optimization problem exists, but mainly for Mahanalobis distance metric learning \cite{zadeh2016geometric,bellet2013survey,wang2014kernel}, scarcely for Riemannian metric learning problems concerning the geodesic distance \cite{buttazzo2004optimal,qiu2024estimating,benmansour2010derivatives} and nearly nothing for the other Riemannian objects $(\Exp_\metric,\parTp_\metric,\vol_\metric)$ \cite[Chapter 4-5]{louis2019computational}.
Theoretical results on the optimization related to these objects would greatly enhance the understanding of the topic.

A deeper understanding of the space of Riemannian metrics \cite{ebin1968space} is essential for constructing relevant parameterizations \cite{tuschmann2015moduli} and for advancing the design of general priors.
Moreover, the geometric structure of these spaces plays a crucial role in the design of effective optimization trajectories, underscoring the importance of integrating geometry into future methodological developments.

\paragraph{Geometry-Aware Optimization Methods}
Many Riemannian optimization methods \cite{sato2021riemannian,kochurov2020geoopt} have been developed to account for the geometric structure of the space associated with the learnable parameters.
Incorporating geometry into optimization can be compared to employing a Newton method rather than a first-order gradient descent as it leverages additional information about the underlying manifold.

However, these methods rely on the Riemannian manifolds being sufficiently simple to facilitate the implementation of key mathematical operations, such as projections onto the tangent space, the exponential map, and parallel transport.
To the best of our knowledge, these operations are not yet widely available, cheap, or standardized for the space of Riemannian metrics \cite{ebin1968space,gilmedrano1991riemannian}.
Further advancements on the choice of the metric on $\mathcal{G}_\manifold$ are anticipated to address these limitations, in the same way that the Log-Euclidean metric \cite{arsigny2007geometric} improves the numerical tractability over the affine-invariant metric while still offering a useful structure.

\subsection{Applied Research Avenues}
Numerical and modeling challenges are central to achieve the widespread adoption of Riemannian metric learning, paving the way for more applied research directions.

\paragraph{Enhanced Software and Methodological Tools}
Several libraries currently exist for statistics and optimization on Riemannian manifolds, such as Geomstats \cite{miolane2020geomstats} and Geoopt \cite{kochurov2020geoopt}, as well as libraries for distance metric learning, including pydml \cite{suarez2020pydml} and PyTorch Metric Learning \cite{musgrave2020pytorch}, particularly in the context of deep learning.
However, to date, no comprehensive library specifically tailored for Riemannian metric learning has been developed.

This gap is understandable, given the wide diversity of use cases in Riemannian metric learning, which complicates the design of general-purpose algorithms.
Nonetheless, as outlined in \Cref{sec:abstract}, the associated optimization problem is inherently general.
To be more specific, a library specialized in the computation of the Riemannian objects according to the metric parametrization $\mathcal{F}(\metric)=(\dist_\metric,\Exp_\metric,\parTp_\metric,\vol_\metric)$ is to be waited.
It is reasonable to anticipate that future research will focus on making these methods scalable (\Cref{sec:computing}) akin to advancements observed in deep learning.

\paragraph{Discovering New Applications}
As specialized and user-friendly libraries for Riemannian metric learning become available, the range of potential applications will naturally expand.
Moreover, as the conceptual framework of Riemannian geometry becomes better understood, practitioners are likely to identify relevant applications in their respective fields.
In particular, the Ricci curvature remains insufficiently explored and utilized, as highlighted in \cite{li2023geometry,farzam2024geometry}.
Since the curvature of loss functions serves as an indicator of a model's generalization capacity \cite{thomas2020interplay}, Riemannian metric learning could contribute to improvements in both optimization and generalization, much as it has already proven beneficial for sampling \cite{cui2024optimal}.

\section{Conclusion}
\label{sec:conclusion}

All in all, this paper has highlighted the richness and potential of the Riemannian metric learning framework across various applications (\Cref{sec:applications}).
As we have seen, distance metric learning has helped to increase noticeably the performance of classification methods in the past. 
We begin to see similar effects of \emph{Riemannian} metric learning in a range of applications, improving the performance in, \eg, generative modeling, trajectory inference, and causal inference.
Since we have argued that tools descending from a Riemannian metric are applicable in many contexts in a flexible manner, we anticipate this list of successful applications will grow further as more researchers discover the possibilities of Riemannian metric learning.
Thus, we hope that his review provides them with an accessible and motivating introduction and we encourage readers to explore new ways to leverage it in their respective fields.

Despite its promise, Riemannian metric learning still faces significant computational and theoretical challenges (\Cref{sec:future}). 
While there is substantial theoretical knowledge about the infinite-dimensional Riemannian manifold of Riemannian metrics, it is not connected to any statistical theory or similar for the metric learning problem.
From a methodological perspective, research in distance metric learning is already well-established, providing a solid foundation upon which Riemannian metric learning could build (\Cref{sec:distance_metric_learning,sec:objectives_classification}).
Providing a solid theoretical framework along with flexible computational tools will undoubtedly be demanding, but also offers exciting opportunities to further the adoption and impact.


In conclusion, we firmly believe that Riemannian metric learning represents an exciting new frontier in representation learning, with substantial potential for both theoretical advancements and practical applications. 
However, realizing this potential requires addressing a number of computational and theoretical challenges to ensure scalable methods and rigorous guarantees.

\clearpage

\bibliography{ref.bib}

\clearpage

\appendix

\section{Some Rigorous Aspects of Riemannian Geometry}
\label{sec:appendix_extended_preli}

\paragraph{Manifold}
We begin with the basic definitions introducing our objects of study.
\begin{definition}[Differentiable manifold]
    \label{def:differentiable_manifold}
    A \emph{differentiable manifold} \(\manifold\) of dimension \(d < \infty\) is a set together with a family of injective maps \(\chart_\alpha \colon U_\alpha \subset \R^d \to \manifold\) of open sets \(U_\alpha\) of \(\R^d\) into \(\manifold\) such that
    \begin{enumerate}
        \item \( \bigcup_\alpha \chart_\alpha(U_\alpha) = \manifold \), and
        \item for any pair \(\alpha, \beta\) with \(\chart_\alpha(U_\alpha) \cap \chart_\beta(U_\beta) = W \neq \emptyset\), the sets \(\chart_\alpha^{-1}(W)\) and \(\chart_\beta^{-1}(W)\) are open in \(\R^d\) and the map \(\chart_\beta^{-1} \circ \chart_\alpha\) is differentiable.
    \end{enumerate}
    The pair \((U_\alpha, \chart_\alpha)\) with \(p \in \chart_\alpha(U_\alpha)\) is called a \emph{parametrization}  of \(\manifold\) at \(p\).
\end{definition}

Typically one assumes that the differentiable structure in \Cref{def:differentiable_manifold} is maximal with respect to the given conditions.
This is can be achieved by extending the structure by parametrizations compatible with condition (2).

\paragraph*{Definition of Covariant Derivative}

\begin{definition}
	\label{def:covariant_derivative}
Let \( \manifold \) be a smooth manifold, \( \vf \) a smooth vector field on \( \manifold \), and \( \wf\).
 The \emph{covariant derivative} of \( \wf \) with respect to \( \vf \), denoted by \( \nabla_\vf \wf \), is defined as a map:
\[
\nabla\colon \vecFields \times \vecFields \to \vecFields,
\]
where \( \vecFields \) is the space of smooth vector fields on \( \manifold \). A covariant derivative satisfies the following properties:

\begin{enumerate}
    \item \textbf{Linearity:} For any smooth vector fields \( Z,\vf,\wf\in \vecFields \) and $\alpha\in \R$,
    \[
    \nabla_Z (\vf+\alpha \wf) = \nabla_Z \vf + \alpha \nabla_Z \wf.
    \]

    \item \textbf{Product Rule:} For any vector field \(\vf, \wf\in \vecFields \) and smooth function \( f\colon\manifold\to \R \),
    \[
    \nabla_\vf (f \wf) = (\nabla_\vf f) \wf + f (\nabla_\vf \wf).
    \]
	where $\nabla_\vf f=\nabla f^\top \vf$ where $\nabla$ at right denotes the usual gradient operator.

    \item \textbf{$C^1$-linearity:} For any smooth function $f,g\colon\manifold \to \R$ and smooth vector fields $Z,\vf,\wf\in \vecFields$,
    \[
    \nabla_{fZ+g\wf} \vf  = f\nabla_{Z} \vf+g\nabla_{\wf} \vf  .
    \]
	
\end{enumerate}
\end{definition}

\paragraph*{General definition of covariate derivative in infinite dimension.}

One can obtain a coordinate-free formulation, which is also useful for infinite dimensional manifolds, by defining a bilinear operator through \(\Gamma(\partial_{i}\chart, \partial_{j}\chart) = \sum_{k=1}^d \Gamma^k_{ij} \partial_{k}\chart \)
and realizing that because of \Cref{prop:christoffel} this is the unique bilinear form fulfilling the property of the following definition.
\begin{definition}[Christoffel operator]
   	\label{def:christoffel_operator}
   	For \(p \in \manifold\)  the Christoffel operator \(\Gamma = \Gamma_p\) is a mapping \(\Gamma_p\colon T_p\manifold \times T_p\manifold \rightarrow T_p\manifold\).
   	For \(U,V \in T_\point\manifold\) the evaluation \(\Gamma_p(U,V)\) is defined implicitly by 
   	\begin{align}
       	\metric_p(\Gamma_p(U,V),W) = \frac{1}{2} \Big( \left(D_p \metric\right)(V)(U,W) + \left(D_p \metric\right)(U)(V,W) - \left(D_p \metric\right)(W)(U,V) \Big) \, .
    \end{align}
   	for all \( W \in T_\point\manifold\)
\end{definition}

Next, we will use this operator to define a derivative of vector fields.
To this end, let \(\curve \colon [0,1] \to \manifold\) be a smooth curve, \(W\colon[0,1] \to T\manifold\) a vector field along the curve, i.e.\ defined through \(W(t) = \sum_{l=1}^d w_l(t) \partial_l\chart(\curve(t)) \in T_{\curve(t)}\manifold\) for parameters \(w_l\), and lastly let \(\dot W(t) \coloneqq \sum_{l=1}^d \dot w_l(t)\partial_l\chart(\curve(t))\) also define a vector field along \(\curve\).
Then, by projecting the derivative of \(W(t)\) onto the tangent space, i.e.\ ignoring the normal components in \eqref{eq:param_2nd_deriv}, we get the following definition (stated again in a coordinate-free fashion).
\begin{definition}[Covariant derivative]
   	Let \(\gamma \colon I \rightarrow \manifold \) be a curve and \(W \colon I \rightarrow T\manifold \) a vector field  along \(\gamma\).
   	We define the covariant derivative \(\covDeriv W\) of \(W\) along \(\gamma\) at \(p = \gamma(t)\)  for \(t \in I\)  by
   	\begin{align}
      		\label{eq:covariant_derivative}
      		\metric_{\gamma(t)} \Big( \covDeriv W(t), U \Big) = \metric_{\gamma(t)}  \Big( \dot W(t) + \Gamma_\point\left(W(t), \dot \gamma(t)\right), U \Big)\, \quad \forall \, U \in T_{\gamma(t)}\manifold \, .
    \end{align}
\end{definition}
This covariant derivative is a generalization of the usual Euclidean directional derivative of a vector-valued function to Riemannian manifolds.
Indeed, for Riemannian manifolds that are isometrically embedded into Euclidean space (\eg~the case of embedded surface illustrated before) it is exactly the projection of the Euclidean directional derivative onto the tangent space.

\end{document}